\documentclass[10pt,twocolumn,letterpaper]{article}

\usepackage{cvpr}
\usepackage{times}
\usepackage{epsfig}
\usepackage{graphicx}
\usepackage{amsmath}
\usepackage{amssymb}
\usepackage{xcolor,colortbl}
\usepackage{booktabs}
\usepackage{pifont}
\usepackage{subfiles}
\usepackage{csquotes}
\usepackage{multirow}
\usepackage{subcaption}

\usepackage{url}            %
\usepackage{booktabs}       %
\usepackage{amsfonts}       %
\usepackage{nicefrac}       %
\usepackage{microtype}      %
\usepackage{csquotes}
\usepackage{array}
\usepackage{adjustbox}
\usepackage{xspace}
\usepackage{multirow}
\usepackage{wrapfig}
\usepackage{import}
\usepackage{authblk}

\usepackage{diagbox}
\usepackage[font=small,skip=5pt]{caption}
\usepackage{makecell}
\usepackage{enumitem}
\usepackage{gensymb}
\usepackage{wrapfig}
\usepackage{ragged2e}
\usepackage{algorithm,algpseudocode}
\usepackage{color, colortbl}
\usepackage{bbm}
\definecolor{Gray}{gray}{0.95}

\def\ours{\texttt{\textbf{VALHALLA}}\xspace}
\def\oursm{\texttt{\textbf{VALHALLA(M)}}\xspace}

\DeclareMathOperator*{\argmax}{arg\,max}
\DeclareMathOperator*{\argmin}{arg\,min}

\makeatletter
\renewcommand\AB@affilsepx{, \protect\Affilfont}
\makeatother

\usepackage[pagebackref=true,breaklinks=true,letterpaper=true,colorlinks,bookmarks=false]{hyperref}

\cvprfinalcopy % *** Uncomment this line for the final submission

\def\cvprPaperID{9502} % *** Enter the CVPR Paper ID here

\ifcvprfinal\fi
\begin{document}

%%%%%%%%% TITLE
\title{\ours: Visual Hallucination for Machine Translation}

\author[1]{Yi Li\thanks{Work done during an internship at the MIT-IBM Watson AI Lab.}}
\author[2]{Rameswar Panda}
\author[3]{Yoon Kim}
\author[2]{Chun-Fu (Richard) Chen}
% \author[2,4]{Chun-Fu (Richard) Chen}
\author[2]{\\Rogerio Feris}
\author[2]{David Cox}
\author[1]{Nuno Vasconcelos}
\affil[1]{UC San Diego}
\affil[2]{MIT-IBM Watson AI Lab}
\affil[3]{MIT CSAIL}
% \affil[4]{JPMorgan Chase, FLARE}

% \and

\maketitle
\thispagestyle{empty}

%%%%%%%%% ABSTRACT 
\begin{abstract}
Designing better machine translation systems by considering auxiliary inputs such as images has attracted much attention in recent years. While existing methods show promising performance over the conventional text-only translation systems, they typically require paired text and image as input during inference, which limits their applicability to  real-world scenarios. In this paper, we introduce a visual hallucination framework, called VALHALLA, which requires only source sentences at inference time and instead uses hallucinated visual representations for multimodal machine translation. In particular, given a source sentence an autoregressive hallucination transformer is used to predict a discrete visual representation from the input text, and the combined text and hallucinated representations are utilized to obtain the target translation. We train the hallucination transformer jointly with the translation transformer using standard backpropagation with cross-entropy losses while being guided by an additional loss that encourages consistency between predictions using either ground-truth or hallucinated visual representations. Extensive experiments on three standard translation datasets with a diverse set of language pairs demonstrate the effectiveness of our approach over both text-only baselines and state-of-the-art methods. Project page: \url{http://www.svcl.ucsd.edu/projects/valhalla}.

\end{abstract} 

\vspace{-4mm}
%%%%%%%%% Introduction %%%%%%%%%
\section{Introduction}
\label{sec:introduction}

Machine Translation (MT) is a core task in natural language processing and has undergone several paradigm shifts over the past few decades, from early rules-based systems~\cite{nyberg-iii-mitamura-1992-kant} to pipelined statistical MT approaches~\cite{koehn-etal-2007-moses,lopez2008statistical} to recent end-to-end neural network-based models ~\cite{cho-etal-2014-learning,Sutskever2014,bahdanau2014neural,vaswani2017attention}. While such advances  have led to impressive results on standard benchmarks, existing systems by and large utilize text-only information and lack any explicit grounding to the real world. There has thus been a growing interest in developing \emph{multimodal} MT systems that can  incorporate rich external information into the modeling process. 

\begin{figure}
    \centering
    \includegraphics[width=\linewidth]{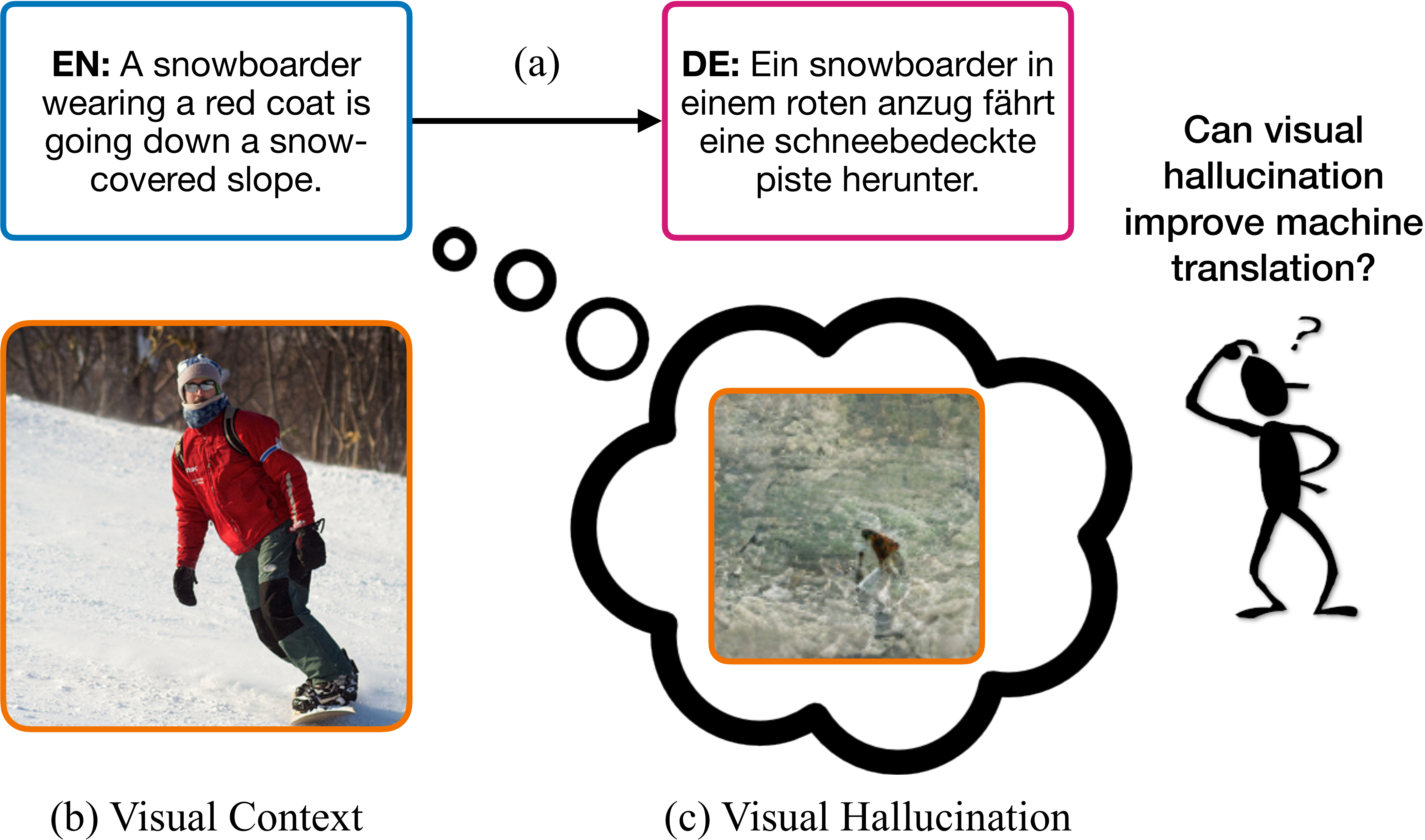} 
    % \vspace{1mm}
    \caption{Visual context such as images has been exploited in designing better machine translation systems. Different from most existing methods that require manually annotated sentence-image pairs as the input during inference, we introduce \ours, that leverages hallucinated visual representation from the source sentences at test time for improved machine translation.}
    \label{fig:teaser}
\end{figure}

Consider the example in Figure~\ref{fig:teaser}(a), where a source sentence in English (blue box) is to be translated to a target sentence in German (red box). Since both sentences depict the same visual scene, shown in Figure~\ref{fig:teaser}(b), \textit{there is common grounding information across the two sentences}.  More generally, while there are many different ways to describe a situation in the physical world, the underlying visual perception is shared among speakers of different languages.
The addition of visual context in the form of images is thus likely to help the machine translation. In particular, grounding should improve the data-efficiency of translation methods and benefit translation in low resource scenarios.

This has motivated much recent work on vision-based multimodal machine translation (MMT), which aims to improve machine translation systems by utilizing the visual modality~\cite{calixto2017doubly,lin2020dynamic,zhou2018visual,ive2019distilling}. These methods typically require source sentences to be paired with the corresponding images during training \emph{and} testing, which hinders their applicability to settings where images are not available during inference.
% Developing MMT systems that do not require images during inference but can still leverage visual information in a useful way thus remains an important open problem.
In this work we consider the question of whether a system that has access to images only at training time can generalize to these settings. 
We hypothesize that \textit{\enquote{visual hallucination, i.e., the ability to imagine visual scenes, can be leveraged to  improve machine translation systems}.} Under this hypothesis, a translation system with access to images at training time could be taught to abstract an image or visual representation of the text sentence, as shown in Figure~\ref{fig:teaser}(c), in order to ground the translation process. At test time, this \emph{abstracted} visual representation could be used in lieu of an actual image to perform multimodal translation.
%Rather than using a \textit{real image} that is either manually paired or retrieved during inference, we could instead exploit an \textit{abstract image or visual representation} depicting similar scenes, as shown in Figure~\ref{fig:teaser}(c)---if only we knew how to synthesize and integrate such representation effectively in the machine translation systems.

We introduce a simple yet effective \textbf{V}isu\textbf{AL} \textbf{HALL}ucin\textbf{A}tion (\ours) framework, which incorporates images at training time to produce a more effective text-only model for machine translation. As is usual for  machine translation, the goal is to train a model that only sees source sentences at test time. However, during training, the model is trained to complement the text representation extracted from the source sentence with a latent visual representation that mirrors the one extracted from a real image (paired with the source sentence) by an MMT system. We achieve this by training an autoregressive hallucination transformer over a discrete codebook (learned using  VQGAN-VAE~\cite{esser2021taming}) to predict visual tokens from the input source sentences for multimodal translation.  

\ours consists of a pair of transformers: a visual hallucination transformer that maps the source sentence into a discrete image representation, and an MMT transformer that maps the source sentence paired with its discrete image representation into the target sentence. 
% Both transformers are trained jointly, with a combination of hallucination, translation, and consistency losses, which ensures the reliable performance of the visual hallucination transformer at inference time.
% guarantee {\color{red} NEEDS MORE WORK
% formulate training losses in a way that focuses on translation performance
% We train our framework using standard cross-entropy losses including consistency between predictions using either ground-truth or hallucinated representation for reliable performance of the visual hallucination transformer at inference time.
We train the transformer models end-to-end with a combination of hallucination, translation, and consistency losses.
% to ensure the reliable performance of the visual hallucination transformer at inference time.
% While it looks trivial, end-to-end optimization is challenging 
As sampling of the discrete image representations (i.e., visual hallucinations)
% prevents loss gradients from backpropagating through the module. To address this, 
is non-differentiable, we rely on a Gumbel-Softmax relaxation~\cite{jang2016categorical,maddison2016concrete} to effectively train the hallucination transformer jointly with the translation transformer.
To the best of our knowledge, ours is the first  work that successfully
leverages an  autoregressive image transformer jointly with the translation transformer to hallucinate discrete visual representations. We find that discrete visual representations lead to improved performance compared to continuous visual embeddings used in existing MMT methods~\cite{wu2021good,lin2020dynamic,yao2020multimodal,zhang2020neural,long2021generative}.
% extracted using a ResNet~\cite{he2016deep} pretrained on external datasets like ImageNet~\cite{krizhevsky2012imagenet} or COCO~\cite{lin2014microsoft}.

Extensive experiments on three standard MT datasets (Multi30K~\cite{elliott2016multi30k}, WIT~\cite{srinivasan2021wit} and WMT~\cite{bojar2014findings}) with a diverse set of language pairs and different scales of training data (in total $13$ pairs) demonstrate the superiority of \ours over strong translation baselines. 
\ours yields an average $2$$\sim$$3$ BLEU improvement over the text-only translation baseline, while consistently outperforming the most relevant state-of-the-art MMT methods that make use of continuous image representations~\cite{zhang2020neural,long2021generative}. The gains over the text-only baseline are as large as $+3.1$ BLEU on under-resourced translation settings, such as the EN$\rightarrow$RO and EN$\rightarrow$AF tasks from WIT, confirming the hypothesis that visual hallucinations can have significant practical value in these settings. This is also confirmed by additional analysis suggesting that, under limited textual context, \ours models do leverage visual hallucination to generate better translations.

% \vspace{-2mm}
%%%%%%%%% Related Work %%%%%%%%%
% \vspace{-4mm}
\section{Related Work}
\label{sec:relatedwork}

% \paragraph{Machine Translation.} Much progress has been made in developing a variety of ways to translate sentences across a pair of languages such as statistical methods~\cite{lopez2008statistical} in the early years to neural machine translation (NMT) with Seq2Seq architectures~\cite{sutskever2014sequence}. Many successful NMT architectures are usually based on encoder-decoder framework with attention~\cite{bahdanau2014neural}, e.g., transformer~\cite{vaswani2017attention} which achieves state-of-the-art translation quality compared to recurrent or convolutional networks.   
% With the development of NMT, multilingual neural machine translation has attracted a great amount of attention in the recent years~\cite{johnson2017google,aharoni2019massively,zhu2021serial}.
% Developing unsupervised or semi-supervised NMT methods to alleviate the need of well-annotated large parallel corpora is also another popular trend in neural machine translation~\cite{artetxe2017unsupervised,lample2017unsupervised,lample2018phrase,cheng2019semi}. 

% \vspace{1mm}
\noindent\textbf{Multimodal Machine Translation.} 
% Many multimodal machine translation (MMT) methods have been recently proposed with the goal of improving conventional text-only translation systems by utilizing visual modality~\cite{specia2016shared,wang2019vatex,calixto2017doubly,lin2020dynamic,zhou2018visual,ive2019distilling}. 
% MMT has been studied from multiple perspectives~\cite{specia2016shared,wang2019vatex,calixto2017doubly,lin2020dynamic,zhou2018visual,ive2019distilling,yin2020novel,yao2020multimodal,liu2021gumbel,caglayan2021cross}. 
MMT has been studied from multiple perspectives~\cite{specia2016shared,wang2019vatex,calixto2017doubly,zhou2018visual,ive2019distilling,yin2020novel,yao2020multimodal,liu2021gumbel,caglayan2021cross,yang2020using,yang2021backpropagation}. 
% Unlike the problem considered in this paper,
Different from our work, 
a few methods~\cite{sigurdsson2020visual,suris2020globetrotter} use visual alignment for unsupervised word mapping and translation by retrieval.
% While DCCN~\cite{lin2020dynamic} introduces a capsule network to effectively capture visual features at different granularities, GMNMT~\cite{yin2020novel} explores multimodal graph neural network to exploit interactions among multimodal semantic units for MMT.
% Yao and Yan~\cite{yao2020multimodal} propose multimodal self-attention in Transformer to avoid encoding irrelevant information in images. Gumbel-Attention~\cite{liu2021gumbel} uses a differentiable method to select the visual information and automatically removes the useless parts of the features. 
Unsupervised MMT methods have been proposed in~\cite{su2019unsupervised,huang2020unsupervised}. Recent works show that visual context does not help translation reliably~\cite{elliott2017findings,wu2021good} or is mostly beneficial under limited textual context~\cite{caglayan2019probing,elliott2018adversarial}. Most MMT methods assume images as input at test time, which hinders their potential applications.
Most relevant to our approach are UVR-NMT~\cite{zhang2020neural} and ImagiT~\cite{long2021generative}. 
UVR-NMT uses a token-to-image lookup table to improve text-only NMT but requires retrieval of images during inference to match source language keywords. ImagiT uses a generative adversarial model to synthesize \textit{continuous} image features for MMT. This differs from \ours, which uses a hallucination model to predict \textit{discrete} visual tokens from input text. In addition, ImagiT requires a computationally-heavy image captioning module, while our approach offers more flexible visual hallucination by using a transformer that autoregressively models text and image tokens as a single data stream. 
% \ours also avoids the usual instability pitfalls associated with adversarial learning~\cite{long2021generative}.
% Our experiments quantitatively compare the two approaches, showing that our approach performs significantly better in machine translation across a diverse set of languages.
% i.e., leveraging visual hallucinations for improving more general supervised machine translation, especially text-only NMT. 

% \todo{Vision-language pretraining~\cite{caglayan2019probing,caglayan2021cross}\dots}

\begin{figure*}
    \centering
    \includegraphics[width=\linewidth]{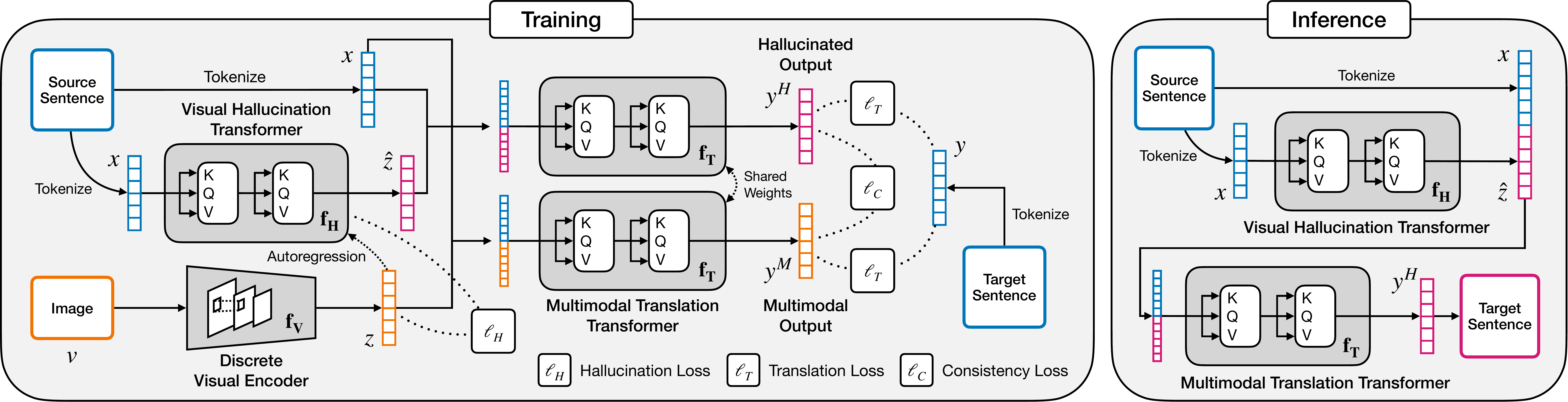}
    \caption{Overview of \ours Architecture for Machine Translation. \textbf{Left}: Training pipeline of \ours. Translation outputs are gathered from two streams of input, either with ground-truth visual tokens $z$ or hallucinated representation $\hat{z}$, and optimized on a combination of \textit{hallucination}, \textit{translation} and \textit{consistency} losses. \textbf{Right}: Inference process of \ours in the absence of visual inputs.
    % \rp{tilde should be hat in hallucination tokens?} 
    % \rc{In the training, what is the dash line between $\phi_V$ and Visual Hallucination Transformer? And a minor thing, at the training diagram, we called Visual Hallucination Transformer but Visual Hallucination in the inference. Should we make it consistent? same for multimodal translation} \yl{Dashed line indicates hallucination transformer $f^H$ is trained augoregressively on $\phi^V$. Inference figure fixed}
    }
    \label{fig:framework} \vspace{-1mm}
\end{figure*}

\vspace{0.5mm}
% \noindent\textbf{Visually-Grounded Language Learning.} 
\noindent\textbf{Vision-Language Learning.} 
Visual grounding 
% is becoming increasingly attractive due to its potential 
has been used to improve performance and data-efficiency across many tasks~\cite{sileo2021visual,norlund2021transferring}, such as semantic parsing~\cite{shi2019visually}, co-reference resolution~\cite{kong2014you}, representation learning~\cite{bordes2020incorporating,kiela2017learning,singhal2019learning}, grammar induction~\cite{shi2019visual,zhao2020pcfg,jin-schuler-2020-grounded,hong2021vlgrammar,zhang2021grammar}, lexicon learning~\cite{vulic2016multi}, and language learning with multimodal knowledge distillation~\cite{tang2021vidlankd}, or mapping language tokens with images~\cite{tan2020vokenization}. 
Conversely, image-text correspondence has also been exploited to improve vision tasks, such as image retrieval~\cite{pereira2014cross} and classification~\cite{radford2021learning}.
Despite recent progress, improving machine translation with no visual input at test time remains a challenging and largely under-addressed problem.

\vspace{0.5mm}
\noindent\textbf{Text-to-Image Generation.} Generating images from text has been extensively studied~\cite{ramesh2021zero,esser2021taming,reed2016generative,nguyen2017plug}.
% with the successes of VAEs~\cite{kingma2013auto} and GANs~\cite{goodfellow2014generative}. 
Representative works use
GANs~\cite{reed2016generative,xu2018attngan,zhang2017stackgan,qiao2019mirrorgan,zhu2019dm,zhang2021cross} to synthesize photo-realistic scenes with high semantic fidelity to their conditioned text descriptions. 
% Recently, XMCGAN~\cite{zhang2021cross} utilizes a GAN with contrastive losses, 
% DM-GAN~\cite{} proposes a GAN model combined with a dynamic memory component to generate high-quality images.
% while 
DALL-E~\cite{ramesh2021zero} proposes an autoregressive transformer with discrete VAEs~\cite{oord2017neural} to create images from text for a wide range of concepts expressible in natural language. While our approach is inspired by these works, the goal of the present work is to hallucinate discrete visual representations for improving machine translation instead of generating high-quality photo-realistic images.

\vspace{0.5mm}
\noindent\textbf{Modality Hallucination.} \ours is also related to prior work on learning using side information~\cite{vapnik2009new}.
% , where additional information or modality is only available during training time. 
% At test time, the model either operates without the side information or hallucinates the missing information using an additional network. 
A model to hallucinate depth features from RGB input for object detection 
% task 
is proposed in~\cite{hoffman2016learning}.
Graph distillation has been used to transfer
multimodal privileged information across domains for action detection~\cite{luo2018graph}. Garcia et al.,~\cite{garcia2018modality} propose modality distillation for video action recognition.

% \vspace{-2mm}
%%%%%%%%% Proposed Method %%%%%%%%%
\section{Proposed Method}
\label{sec:method}
% \vspace{-1mm}
% In this section, we delineate the \ours architecture  for improved machine translation, its training and inference procedures, and loss functions used to optimize each module.

% \input{figures/mmt}
% \input{figures/framework}

% In this section, we describe the proposed \ours framework. 
Given a corpus of source sentence $x \in \mathcal{X}$ and visual context $v \in \mathcal{V}$, typically images, our goal is to train a machine translation system that can translate a source sentence $x$ into a sentence $y \in \mathcal{Y}$ in a target language  without requiring images at inference time. 
% the system is trained to predict sentence $y^T \in \mathcal{Y}^T$ in the target language. At inference time, the model has to perform text-only translation on $x^S$ in absence of visual data $v$.
% \todo{Update figures}

\subsection{Preliminaries}
\label{sec:method_prelim}
% \vspace{-1mm}
\noindent
\textbf{Machine Translation.} Contemporary MT systems are generally based on the encoder-decoder framework with attention \cite{bahdanau2014neural,vaswani2017attention}. 
% which achieves state-of-the-art translation quality.
% compared to recurrent or convolutional networks.
% (MT) is the task of predicting sentences $x^T$ in a target language from source language sentences $x^S$.
% \textbf{Transformers}~\cite{vaswani2017attention} are sequence-to-sequence encoder-decoder models trained on 
Given sequence pairs $(x, y)$, where $x = \left(x_1, \dots, x_{S}\right)$ is the source sentence of length $S$ and $y = \left(y_1, \dots, y_{T}\right)$ is the target sentence of length $T$,
% \begin{eqnarray}
%     & x = \left(\phi_1^S, \dots, \phi_{l_S}^S\right) \in \{1, \dots, N_S\}^{l_S}, \\
%     & y = \left(\phi_1^T, \dots, \phi_{l_T}^T\right) \in \{1, \dots, N_T\}^{l_T}
% \end{eqnarray}
% denote tokenized text sequences of length $l_S$ and $l_T$ using dictionaries of size $N_S$ and $N_T$, respectively, 
a transformer $\mathbf{f}_{\mathbf{T}} = (\mathbf{f}_{\mathbf{T}}^{\textrm{enc}}, \mathbf{f}_{\mathbf{T}}^{\textrm{dec}})$ models the likelihood of target tokens conditioned on the input sequence as
% \vspace{-2mm}
\begin{equation}
\begin{aligned}
    p\left(y \mid x; \mathbf{f}_{\mathbf{T}}\right) &= \prod_{i=1}^{T} \mathbf{f}_{\mathbf{T}}\left(y_i \mid y_{<i}, x\right) \\
    &\triangleq \prod_{i=1}^{T} \mathbf{f}_{\mathbf{T}}^{\textrm{dec}}\left(y_i \mid y_{<i}, \mathbf{f}_{\mathbf{T}}^{\textrm{enc}}(x)\right),
\end{aligned}
\end{equation}
where the decoder $\mathbf{f}_{\mathbf{T}}^{\textrm{dec}}$ predicts probability of output tokens at each location $i$ by attending to encoder output  $\mathbf{f}_{\mathbf{T}}^{\textrm{enc}}(x)$ and previous target tokens $y_{<i}$ using a cascade of attention layers. $\mathbf{f}_{\mathbf{T}}$ is trained by minimizing the cross-entropy loss
\begin{equation}
    \ell_T(\mathbf{f}_{\mathbf{T}}) = \mathbb{E}_{(x, y)}\left[-\log p\left(y \mid x; \mathbf{f}_{\mathbf{T}}\right)\right].
\end{equation}

\noindent
\textbf{Multimodal Machine Translation.} MMT considers a visual input $v$ as a complementary information source for machine translation. 
% (MMT) augments MT by considering visual inputs $v$, typically images, as additional modality that complements the textual information. 
MMT systems typically use an encoder $\mathbf{f}_{\mathbf{V}}$ to map an image into a latent visual representation $z = \mathbf{f}_{\mathbf{V}}(v)$, which are fed into a modified decoder (e.g., by concatenating $z$ with the word embeddings of $x$) to obtain the probabilities conditioned on visual input,
\begin{equation}
    p\left(y \mid x, z; \mathbf{f}_{\mathbf{T}}\right) = \prod_{i=1}^{T} \mathbf{f}_{\mathbf{T}}\left(y_i \mid y_{<i}, x, z\right).
\end{equation}
% {\color{red} SHOULD f have index i here?}
MMT models are trained on a dataset of triplets $(x, v, y)$ by optimizing a translation loss based on cross-entropy
\begin{equation}
    \ell_T(\mathbf{f}_{\mathbf{T}}; z) = \mathbb{E}_{(x, z, y)}\left[-\log p\left(y \mid x, z; \mathbf{f}_{\mathbf{T}}\right)\right].
    \label{eqn:mmt_loss}
\end{equation}
While incorporating visual information improves the translation performance of MMT systems over their text-only counterparts, it requires sentence-image pairs as input at \emph{inference} time. This greatly limits the application of MMT systems in real world scenarios. We next introduce our \ours framework, which addresses this constraint using discrete visual embedding and a hallucination module that predicts visual tokens from textual input for text-only translation.
% \vspace{-1mm}
\subsection{Approach Overview}
% \vspace{-1mm}
The overall \ours framework is illustrated in Figure~\ref{fig:framework}. The architecture consists of three neural network modules: A \textbf{discrete visual encoder} $\mathbf{f}_{\mathbf{V}}$ for mapping input images into sequences of discrete tokens; a \textbf{hallucination transformer} $\mathbf{f}_{\mathbf{H}}$ that predicts visual representations from the source sentence; and a \textbf{multimodal translation transformer} $\mathbf{f}_{\mathbf{T}}$ that predicts the target sentence from the concatenated sequence of text and visual tokens. 

During training, where input sentence-image pairs $(x, v)$ are available, the translation output is predicted through two streams: \emph{Multimodal} (bottom of Figure~\ref{fig:framework}) and \emph{hallucination} (top). 
% The former uses hallucinated visual representations $z$, while the latter uses ground-truth tokens $z$ extracted from the input image. 
The former uses ground-truth (discrete) visual representations $z$ extracted from the input image, while the latter uses \emph{hallucinated} representations $\hat{z}$. 
This produces two distributions $y^M$ and $y^H$ respectively, which are trained against the target sequence $y$ with the cross entropy loss. Training losses also encourage consistency between predictions using either ground-truth or hallucinated visual representations, which is necessary for reliable performance of the visual hallucination module at inference time. As images associated with source sentences are not available at test time, the model utilizes the hallucination stream to generate pseudo-visual tokens and subsequently the translation output, conditioned on the unimodal text input $x$ alone.

% In the following subsections, we detail the design and training of \ours modules. Section~\ref{sec:method_dvae} describes the discrete visual encoder model $\mathbf{f}_{\mathbf{V}}$ and its pretraining procedure. Section~\ref{sec:method_halluc} introduces the visual hallucination transformer $f^H$, and associated losses. Section~\ref{sec:method_mmt} discusses the design of MMT transformer $f^T$ and finally section~\ref{sec:method_train} presents the joint optimization of both transformers using the Gumbel-softmax method. 
%\rp{We can remove this depending on space}
% \vspace{-1mm}
\subsection{Discrete Visual Encoding}
\label{sec:method_dvae}
% \vspace{-1mm}
MMT is typically implemented by combining text input with \emph{continuous} visual embeddings, such as convolutional features extracted from a pretrained ResNet~\cite{he2016deep}. In this work, we instead explore the use of a \emph{discrete} visual encoder~\cite{oord2017neural,razavi2019generating,ramesh2021zero,esser2021taming}. This has two key advantages over a continuous embedding. First, images embedded into a sequence of discrete tokens can be easily concatenated with textual inputs (discrete word embeddings) into a multimodal sequence, which can then be processed by a single universal transformer to produce translation outputs. This vision-language fusion is nontrivial under continuous image representations, where complicated aggregation modules have been proposed for both MMT~\cite{wu2021good,yao2020multimodal,lin2020dynamic,yin2020novel} and other vision-language tasks~\cite{chen2021learning,wu2019unified,li2019visual}. Second, while regressing continuous visual representations requires careful design of losses and training schedule to prevent model predictions from collapsing to the mean value, visual hallucination in the discrete space reduces to a sequence-to-sequence learning problem trainable with a vanilla cross-entropy loss~\cite{ramesh2021zero}.

Motivated by this, we use discrete visual token sequences, which are essentially raster-scanned vector quantization maps of input images with respect to a feature codebook learned from training images. 
% {\color{red} I think it heplps to first give a high-level description of what these are, check that it is correct}. 
We implement vector quantization with the VQGAN VAE model of~\cite{esser2021taming}, using a visual encoder $\mathbf{f}_{\mathbf{V}}$ to map input image $v$ into a token sequence as
\begin{equation}
    z = \mathcal{Q}(\mathbf{f}_{\mathbf{V}}(v); E_V).
    % \in \{1, \dots, N_V\}^{V}.
\end{equation}
Here $z=[z_1, \dots, z_V]$ is a grid of discrete tokens laid out as a sequence where $z_i \in \{1, \dots, K\}$, $E_V = \{e^{(k)}\}_{k=1}^{K}$ are the  $d$-dimensional visual codebook of size $K$, and $\mathcal{Q}$ denotes the quantization function
\begin{equation}
    \mathcal{Q}_i(c; E_V) = \argmin_{k \in \{1, \dots, K\}} \lVert c_i - e^{(k)} \rVert_2
    \label{eqn:vis_decode}
\end{equation}
that maps each spatial location $i \in \{1, \dots, V\}$ of feature array $c = \mathbf{f}_{\mathbf{V}}(v) \in \mathbb{R}^{V \times d}$ into the index of its closest visual code in $E_V$. 
Given a multimodal training set $\mathcal{D} = \{(x, v, y)\}$ where $x$, $y$ denote source and target sentences, the image encoder $\mathbf{f}_{\mathbf{V}}$ is trained on collection of images $\{v\}$ by optimizing a combination of reconstruction loss, vector quantization loss~\cite{oord2017neural}, and GAN adversarial loss. We refer the readers to \cite{esser2021taming} for more implementation details of the VQGAN VAE model.

% Given a multimodal training set $\mathcal{D} = \{(x^S, v, y_T)\}$ where $x^S$, $y_T$ denotes source and target sentences respectively, the VQGAN encoder $\mathbf{f}_{\mathbf{V}}$, decoder \todo{symbol}, along with a GAN discriminator \todo{symbol} and the visual codebook $\mathcal{Z}_V$, are pretrained on the collection of images $\{v\}$ by optimizing the \dots, following~\cite{esser2021taming}:
% \begin{equation}
%     \ell_V(\mathbf{f}_{\mathbf{V}}, \dots, \mathcal{Z}_V) = \mathbb{E}_{x \sim p_\mathcal{D}} \left[  \ell_{\textrm{VQ}}(\dots) + \lambda \max_{f_D} \ell_{\textrm{GAN}}(\dots) \right].
% \end{equation}

Once $\mathbf{f}_{\mathbf{V}}$ is learned, MMT feature aggregation becomes trivial as we can simply extend the input sequence of source tokens $x$ to the translation transformer $\mathbf{f}_{\mathbf{T}}$ with $z$ encoded by~\eqref{eqn:vis_decode} by concatenating the word/visual embeddings.
% \vspace{-1mm}
\subsection{Visual Hallucination}
\label{sec:method_halluc}
% \vspace{-1mm}
During inference, when visual inputs are not available, \ours relies on the visual hallucination module $\mathbf{f}_{\mathbf{H}}$ to predict discrete visual tokens $z$ given input text $x$. 
% Rather than directly predicting conditional probabilities $p(z \mid x)$,% we train a transformer~\cite{vaswani2017attention} to autoregressively model the text and image tokens as a single stream of data. Specifically, 
We follow \cite{ramesh2021zero} and implement an autoregressive transformer that models the concatenation of text and image tokens as
\begin{equation}
\begin{aligned}
    p(x, z; \mathbf{f}_{\mathbf{H}}) & = p(x; \mathbf{f}_{\mathbf{H}}) p(z \mid x; \mathbf{f}_{\mathbf{H}}) \\
    &= \prod_{i=1}^{S} \mathbf{f}_{\mathbf{H}}(x_i \mid x_{<i}) \prod_{j=1}^{V} \mathbf{f}_{\mathbf{H}}(z_j \mid z_{<j}, x).
\end{aligned}
\end{equation}
% The prediction at each time step $i$ is a categorical distribution over all $N_V$ possible visual tokens:
% \begin{eqnarray}
%     & \mathbf{f}_{i, k}^H(\phi_T) \triangleq \mathbf{f}_{\mathbf{H}}([\phi_H]_i = k \mid [\phi_H]_{<i}, \phi_T), \\
%     & 1 \le i \le l_V, \quad 1 \le k \le N_V. \nonumber
%     % & i = 1, \dots, l_V, \quad k = 1, \dots, N_V. \nonumber
% \end{eqnarray}

The hallucination transformer is trained to maximize the joint likelihood of $x$ and $z$ by optimizing the cross-entropy \emph{hallucination loss} 
% \rp{Is this consistent with the figure?}
\begin{equation}
    \ell_H(\mathbf{f}_{\mathbf{H}}) = \mathbb{E}_{(x, z)}\left[-\log p(x, z; \mathbf{f}_{\mathbf{H}})\right].
    \label{eqn:halluc_loss}
\end{equation}
We emphasize that as in \cite{ramesh2021zero} we model the joint $p(x, z ;\mathbf{f}_{\mathbf{H}})$ and not just the conditional $p(z \mid x; \mathbf{f}_{\mathbf{H}})$, which was found to improve the results.

The hallucinated visual sequence $\hat{z}$ is then defined as the most likely token predicted by $\mathbf{f}_{\mathbf{H}}$ at each time step $i$,
\begin{equation}
    \hat{z}_i = \argmax_{k \in \{1, \dots, K\}}
    \mathbf{f}_{\mathbf{H}}(z_i = k \mid z_{<i}, x),
    \label{eqn:halluc_argmax}
\end{equation}
where the conditioning $z_{<i}$ is replaced with hallucinated visual sequence $\hat{z}_{<i}$ at inference time.
% \begin{equation}
%     \hat{\phi}_i^V = \argmax_{k \in \{1, \dots, N_V\}}
%     \mathbf{f}_{\mathbf{H}}(\phi_i^V = k \mid \hat{\phi}_{<i}^V, x),
%     \label{eqn:halluc_argmax_inference}
% \end{equation}
% {{\color{blue}This enables the hallucination transformer to perform autoregressive decoding using source text tokens $x$ only.
% % \begin{equation}
% %     [\phi_H]_i = \argmax_{k \in \{1, \dots, N_V\}} f_{i, k}^H(\phi_T), \quad i = 1, \dots, l_V.
% %     \label{eqn:halluc_argmax}
% % \end{equation}
% % By optimizing the hallucination loss of \eqref{eqn:halluc_loss}, we hypothesize that the hallucinated output $\phi_H$ should resemble ground-truth visual tokens $z$, making it an ideal replacement for visual inputs in multimodal translation. {\color{red} Right now, because I see no connection to $y_T$ I don't see what forces this. In any case, even assuming you have the connection through $y_T$, why not introduce an explicit connection between $\phi_T$ and $z$ using a KL-divergence or something like that? Wouldn't this help calibrate the probabilities?}
% % \todo{Move consistency loss here}
% While the hallucination loss of \eqref{eqn:halluc_loss} encourages similarity between hallucinated and ground-truth tokens, we argue that proximity in visual embedding space does not suffice to guarantee agreement at the output of the translation transformer $\mathbf{f}_{\mathbf{T}}$.}
% {\color{red} 
While this enables the hallucination transformer to perform autoregressive decoding using source text tokens $x$ only, it creates a mismatch between the training and inference, which is reflected in the output of the multimodal translation transformer. To reduce this mismatch, we define a \emph{consistency loss}
% }
% {\color{blue}Thus, we define an additional \emph{consistency loss}}
\begin{equation}
    \ell_C(\mathbf{f}_{\mathbf{H}}, \mathbf{f}_{\mathbf{T}}) = \mathbb{E}_{(x, z, y)} \left[\sum_{i=1}^T \text{KL}[y^M_i \, \Vert \, y^H_i]\right],
    % \ell_C(\mathbf{f}_{\mathbf{T}}) = \mathbb{E}_{x, z, \phi_T} \ell_D(y^H, y^M),
    \label{eqn:halluc_consistency_loss}
\end{equation}
where $y_i^M = p(y_i | x, z, y_{<i}; \mathbf{f}_{\mathbf{T}})$ and $y_i^H = p(y_i | x, \hat{z}, y_{<i}; \mathbf{f}_{\mathbf{T}})$ are the next word distributions from ground-truth visual tokens and hallucinated features respectively, and $\text{KL}[y^M_i \Vert y^H_i]$ is the Kullback-Leibler divergence between the two conditional distributions.
\subsection{Optimization}
% \vspace{-1mm}
\label{sec:method_train}

% While feasible to pretrain the visual hallucination transformer $f^H$ and freeze the module when learning the translation transformer $f^T$, we argue that this strategy is suboptimal, since training loss for $f^H$ is not designed for maximizing semantic information in the predicted visual tokens that assists machine translation. 
A remaining challenge  for the joint optimization of the consistency loss of~\eqref{eqn:halluc_consistency_loss} and the translation loss of~\eqref{eqn:mmt_loss} is that
% However, directly optimizing the system end-to-end is unfeasible, as 
the $\argmax$ operator at the output of visual hallucination module (see~\eqref{eqn:halluc_argmax}) prevents loss gradients from backpropagating through $\mathbf{f}_{\mathbf{H}}$. To address this, we use the Gumbel-softmax relaxation~\cite{maddison2016concrete,jang2016categorical} during training, \ie,
\begin{equation}
    \hat{z}_i = \sum_{k=1}^{K} \frac{\exp((\log \pi_{i,k} + g_k) / \tau)}{\sum_l \exp((\log \pi_{i,l} + g_l) / \tau)} o_k,
    \label{eqn:gumbel}
\end{equation}
where $\tau$ is the temperature of the softmax and $o_k$ is a one-hot vector of length $K$ activated at dimension $k$, $g_1, \dots, g_K \sim \mathrm{Gumbel}(0, 1)$ i.i.d., and 
\begin{equation}
    \pi_{i,k} = \mathbf{f}_{\mathbf{H}}(z_i = k \mid z_{<i}, x).
\end{equation}
% and $g_1, \dots, g_{N_V} \sim \mathrm{Gumbel}(0, 1)$ i.i.d., and $\tau > 0$ denotes the softmax temperature. 
% We adopt a temperature annealing schedule with initial $\tau = 5$ and reduce its value by $0.9995 \times$ each iteration as training proceeds, 
We set $\tau = 5$ as initial value and gradually anneal it down to 0 during training~\cite{jang2016categorical,sun2019adashare}, such that \eqref{eqn:gumbel} converges to a one-hot distribution that resembles the use of \eqref{eqn:halluc_argmax} at inference. 

The overall optimization objective of \ours is finally defined as a weighted sum of translation loss, hallucination loss and consistency losses
\begin{equation}
\begin{aligned}
    \ell(\mathbf{f}_{\mathbf{H}}, \mathbf{f}_{\mathbf{T}}) &= \ell_T(\mathbf{f}_{\mathbf{T}}; z) + \ell_T(\mathbf{f}_{\mathbf{T}}; \hat{z}) \\
    & \qquad + \gamma_H \ell_H(\mathbf{f}_{\mathbf{H}}) + \lambda_C \ell_C(\mathbf{f}_{\mathbf{H}}, \mathbf{f}_{\mathbf{T}}),
\end{aligned}
\label{eqn:valhalla_loss}
\end{equation}
where $\gamma_H$ is a hyperparameter that controls tradeoff between hallucination module $\mathbf{f}_{\mathbf{H}}$ recovering ground-truth visual tokens ($\gamma_H \to \infty$) and extracting semantic information useful for machine translation ($\gamma_H \to 0$), and $\lambda_C$ controls the degree of consistency between translation outputs. 
% Loss weights are determined by grid search on validation data, where we find $\lambda_C = \gamma_H = 0.5$ to work consistently well on all datasets.

Finally, we remark that our proposed approach can be seen as a version of latent variable MT where $z = [z_1,\dots,z_V]$ are discrete latent variables that are grounded (i.e., imbued meaning) by being trained against ``ground-truth'' values of $z$ obtained from the real images.
% Instead of marginalizing over $z$ (i.e. $\log p(y | x) = \log \mathbbm{E}_{p(z|x)}[ p(y | x, z)])$ we optimize a lower bound resulting from Jensen's equality (i.e. $\mathbbm{E}_{p(z|x)}[\log p(y | x, z)]$), where this lower bound is itself approximated via the Gumbel-softmax relaxation.
% With the differentiable relaxation of \eqref{eqn:gumbel}, we jointly optimize the visual hallucination and multimodal translation transformers, $f^H$ and $f^T$, on a weighted sum of their respective losses:
% \begin{equation}
% \begin{aligned}
%     \ell(f^H, f^T) &= \ell_T(f^H, f^T) + \gamma_H \ell_H(f^H) \\
%     &= \ell_{MT}(f^T) + \ell_{HT}(f^H, f^T) \\
%     & \qquad + \lambda_C \ell_{CT}(f^H, f^T) + \gamma_H \ell_H(f^H),
% \end{aligned}
% \label{eqn:valhalla_loss}
% \end{equation}
% where $\gamma_H$ is a balancing hyperparameter that controls the tradeoff between the hallucination module $f^H$ recovering ground-truth visual tokens ($\gamma_H \to \infty$) and extracting semantic information useful for machine translation ($\gamma_H \to 0$). The loss weights are determined by grid search on validation data, where we find $\lambda_C = \gamma_H = 0.5$ to work consistently well on all datasets.

% \vspace{-2mm}
%%%%%%%%% Experiments %%%%%%%%%
\section{Experiments}
\label{sec:experiments}
% \vspace{-1mm}
% We conduct extensive experiments on three standard translation datasets to show that \ours outperforms various baselines including state-of-the-art methods. We also perform qualitative analysis to verify the effectiveness of our visual hallucination framework.

\subsection{Experimental Setup}
% \vspace{-1mm}
% \input{tables/main/datasets} 

% \vspace{0.5mm}
\noindent\textbf{Datasets and Tasks.} We evaluate the performance of \ours using three public datasets: Multi30K~\cite{elliott2016multi30k}, Wikipedia Image Text (WIT)~\cite{srinivasan2021wit} and WMT2014~\cite{bojar2014findings}.
% These datasets present a diversity of challenges in machine translation: Multi30K requires models to learn to aggregate vision-language information from a relatively small number of training samples, while WIT and WMT contains translation tasks with different data scales. WMT additionally focuses on translating news articles, which may not be as readily grounded through visual data (compared to Multi30K and WIT), and thus presents an especially challenging test bed for MMT systems. Table~\ref{tab:datasets} shows an overview of all the datasets and tasks. 
% across different language pairs. 
Multi30K~\cite{elliott2016multi30k} is a widely used MMT dataset, consisting of two multilingual expansions (DE and FR) of Flickr30K~\cite{young2014image} dataset. We follow standard evaluation setup of
~\cite{long2021generative,wu2021good} to report performances on three test splits, Test2016, Test2017 and MSCOCO. 
% of trained models
WIT~\cite{srinivasan2021wit} is a large-scale multilingual dataset created by extracting text-image pairs from Wikipedia articles.
% multiple different texts associated with an image from Wikipedia articles and Wikimedia image links.
As no prior work has studied MT on this dataset, we propose a new benchmark with seven language pairs under three settings, \emph{well-resourced} (EN$\rightarrow$\{DE, ES, FR\}),  \emph{under-resourced} (EN$\rightarrow$RO, EN$\rightarrow$AF), and \emph{non-English} (DE$\rightarrow$ES, ES$\rightarrow$FR) splits. We use reference descriptions
% , i.e., captions which are visible on the wiki page directly below the image, 
to obtain parallel sentence-image pairs. 
% with visual annotations. % \rp{Write one sentence on preprocessing and cleaning? We can include the details in supplemental.}
% We generate ground-truth translation pairs by sampling from images with captions annotated in both source and target languages, discarding noisy samples with extremely large or small source-to-target sentence length ratios. 
% The validation and test splits for the original WIT are not publicly available, so we partition the training data to construct new splits for WIT that we will make publicly available for benchmarking.
Detailed dataset preprocessing and cleaning procedure is provided in appendix.

WMT~\cite{bojar2014findings} is a widely-used text-only translation dataset, and we focus on the popular EN$\rightarrow$DE and EN$\rightarrow$FR tasks. 
We use the standard splits of WMT, and further construct two small sets created by sampling from the original set to investigate the performance of \ours in under-resourced settings. Since WMT does not provide aligned images for training, we use CLIP~\cite{radford2021learning} to retrieve top-5 images from Multi30K or WIT datasets to train our 
% hallucination and translation 
transformers. 
% WMT focuses on translating news articles, which may not be as readily grounded through visual data (compared to Multi30K and WIT), and thus presents an especially challenging test bed for MMT systems
% More details about the datasets can be found in supplementary material.
% Note that WMT focuses on translating news articles, which may not be as readily grounded through visual data (compared to Multi30K and WIT), and thus presents an especially challenging test bed for MMT systems.

\vspace{0.5mm}
\noindent\textbf{Models.} We experiment with different transformer model sizes (\emph{Base}, \emph{Small} and \emph{Tiny}). Experiments on Multi30K use the \emph{Small} and \emph{Tiny} configurations, as smaller models have been shown to work better on this dataset~\cite{wu2021good}. For WIT and WMT tasks, we use the base configuration for the well-resourced tasks, while the small configuration is used for both the under-resourced and non-English tasks. 
See appendix for more detailed configurations.   

\vspace{1mm}
\noindent\textbf{Implementation Details.} All our models are trained in three stages. First, we pretrain the discrete visual encoder $\mathbf{f_V}$ on the collection of images associated with training text; we then pretrain the hallucination transformer $\mathbf{f_H}$ using the loss of \eqref{eqn:halluc_loss}; finally, the translation transformer $\mathbf{f_T}$ is learned jointly with $\mathbf{f_H}$ on the combined loss of \eqref{eqn:valhalla_loss}, with hyperparameters $\lambda_C = \gamma_H = 0.5$ determined by a grid search on validation data.
Optimization is performed using Adam~\cite{kingma2014adam} with an inverse square root learning rate schedule and warm-up steps. During inference we use beam search with a beam size of 5. 
% Please refer to supplementary material for more etails.
% See supplementary material 
% for training and evaluation details for each individual dataset and task. 
% for more implementation details including model configurations. 
% or each dataset and task.

\def\textonly{\texttt{T}}
\def\halluc{\texttt{V}}
\def\hallucgt{\texttt{VM}}

\begin{table*}[ht]
\centering
\resizebox{0.95\linewidth}{!}{
\begin{tabular}{@{}l|c|ccc|c|ccc|c@{}}
\Xhline{2\arrayrulewidth} 
\multirow{2}{*}{\textbf{Method}} &
\multirow{2}{*}{\textbf{Model}} & \multicolumn{4}{c|}{\textbf{EN $\rightarrow$ DE}} & \multicolumn{4}{c}{\textbf{EN $\rightarrow$ FR}} \\
\cline{3-10}
&  & \textbf{Test2016} & \textbf{Test2017} & \textbf{MSCOCO} & \textbf{Average} & \textbf{Test2016} & \textbf{Test2017} & \textbf{MSCOCO} & \textbf{Average} \\
 \Xhline{2\arrayrulewidth} 
Transformer-Base & \textonly  & 32.0 ± 0.9 & 23.3 ± 0.8 & 21.3 ± 0.9   & 25.5 ± 0.9     & 59.7 ± 0.2
          & 52.1 ± 0.1         & 42.4 ± 0.6  & 51.4 ± 0.3    \\ \hline
\multirow{3}{*}{Transformer-Small} & \textonly & 38.2 ± 0.4 & 28.8 ± 0.4         
                & 25.8 ± 0.3   & 30.9 ± 0.4    & 58.4 ± 0.4         
                & 50.9 ± 0.3          & 41.6 ± 0.4  & 50.3 ± 0.4     \\
            & \halluc     & 39.4 ± 0.3         & 31.7 ± 0.2
                    & \textbf{27.9 ± 0.3}   &  33.0 ± 0.3     & \textbf{60.5 ± 0.1}
                    & 52.3 ± 0.7 & 43.1 ± 0.3 & 52.0 ± 0.4 \\
            & \hallucgt     & \textbf{39.6 ± 0.3} & \textbf{31.8 ± 0.2}
                    & \textbf{27.9 ± 0.3} & \textbf{33.1 ± 0.3} & \textbf{60.5 ± 0.2}
                    & \textbf{52.4 ± 0.6} & \textbf{43.4 ± 0.2} & \textbf{52.1 ± 0.3} \\ \hline
\multirow{3}{*}{Transformer-Tiny}  & \textonly  & 39.7 ± 0.3  & 31.7 ± 0.5
                    & 28.4 ± 0.2   & 33.3 ± 0.3    & 60.9 ± 0.5
                    & 53.7 ± 0.4          & 44.4 ± 0.2   & 53.0 ± 0.4     \\
            & \halluc     & \textbf{41.9 ± 0.2}         & \textbf{34.0 ± 0.3}
            & 30.3 ± 0.3 & \textbf{35.4 ± 0.3} & 62.3 ± 0.2
            & \textbf{55.1 ± 0.3}          & \textbf{45.7 ± 0.2}  & \textbf{54.4 ± 0.2}     \\
            & \hallucgt     & \textbf{41.9 ± 0.2} & \textbf{34.0 ± 0.3}
                    & \textbf{30.4 ± 0.4}    &  \textbf{35.4 ± 0.3}  & \textbf{62.4 ± 0.3}
                    & 55.0 ± 0.3 & \textbf{45.7 ± 0.4} & \textbf{54.4 ± 0.3} \\
\Xhline{2\arrayrulewidth}
\end{tabular}}
\caption{\textbf{BLEU score on Multi30K}. \textonly: Baseline text-only transformer; \halluc: \ours model with hallucinated visual representations; \hallucgt: \ours model with ground-truth visual representations. 
% On both tasks, \ours (\texttt{V})  consistently outperforms the text-only baseline while being very competitive with \ours (\texttt{VM}). 
Please refer to appendix for METEOR score comparisons.
}
\label{tab:m30k_ours}
\end{table*}

\begin{table*}[ht]
\centering
\resizebox{1\linewidth}{!}{
\begin{tabular}{@{}l|cccccc|cccccc@{}}
\Xhline{2\arrayrulewidth} 
\multirow{3}{*}{\textbf{Method}} & \multicolumn{6}{c|}{\textbf{EN $\rightarrow$ DE}} & \multicolumn{6}{c}{\textbf{EN $\rightarrow$ FR}} \\
\cline{2-13}
 & \multicolumn{2}{c|}{\textbf{Test2016}} & \multicolumn{2}{c|}{\textbf{Test2017}} & \multicolumn{2}{c|}{\textbf{MSCOCO}} & \multicolumn{2}{c|}{\textbf{Test2016}} & \multicolumn{2}{c|}{\textbf{Test2017}} & \multicolumn{2}{c}{\textbf{MSCOCO}} \\
 \cline{2-13}
& \textbf{BLEU}        & \textbf{METEOR}     & \textbf{BLEU}        & \textbf{METEOR}     & \textbf{BLEU}        & \textbf{METEOR} & \textbf{BLEU}        & \textbf{METEOR}     & \textbf{BLEU}        & \textbf{METEOR}     & \textbf{BLEU}        & \textbf{METEOR} \\    
 \Xhline{2\arrayrulewidth} 
\multicolumn{13}{c}{Multimodal Machine Translation} \\ 
\Xhline{2\arrayrulewidth}
% Fusion-Conv~\cite{caglayan2017lium} & 37.0 & 57.0 & 29.8 & 51.2 & 25.1 & 46.0 & 53.5 & 70.4 & 51.6 & 68.6 & 43.2 & 63.1 \\ 
% Trg-Mul~\cite{caglayan2017lium} & 37.8 & 57.7 & 30.7 & 52.2 & 26.4 & 47.4 & 54.7 & 71.3 & 52.7 & 69.5 & 43.5 & 63.2 \\ 
% VAG-NMT~\cite{zhou2018visual} & -- & -- & 31.6 & 52.2 & 28.3 & 48.0 & -- & -- & 53.8 & 70.3 & 45.0 & 64.7 \\ 
% Deliberation NW~\cite{ive2019distilling} & 38.0 & 55.6 & -- & -- & -- & -- & 59.8 & 74.4 & -- & -- & -- & -- \\ 
% Multimodal~\cite{yao2020multimodal} & 38.7 & 55.7 & -- & -- & -- & -- & -- & -- & -- & -- & -- & -- \\ 
Gumbel-Attention~\cite{liu2021gumbel} & 39.2 & 57.8 & 31.4 & 51.2 & 26.9 & 46.0 & -- & -- & -- & -- & -- & -- \\ 
CAP-ALL~\cite{li2021feature} & 39.6 & 57.5 & 33.0 & 52.2 & 27.6 & 46.4 & 60.1 & 74.3 & 52.8 & 68.6 & 44.3 & 62.6 \\ 
GMNMT~\cite{yin2020novel} & 39.8 & 57.6 & 32.2 & 51.9 & 28.7 & 47.6 & 60.9 & 74.9 & 53.9 & 69.3 & -- & -- \\ 
DCCN~\cite{lin2020dynamic} & 39.7 &56.8 & 31.0 &49.9 & 26.7 &45.7 & 61.2 & 76.4 & 54.3 &70.3 & 45.4 & 65.0 \\
\oursm & \textbf{41.9} & \textbf{68.7} & \textbf{34.0} & \textbf{62.5} & \textbf{30.4} &\textbf{57.2} & \textbf{62.4} &\textbf{81.4} & \textbf{55.0} & \textbf{76.4}& \textbf{45.7} & \textbf{71.0}\\
\hline
\rowcolor{Gray} Gated Fusion~\cite{wu2021good} & 42.0 &67.8 & 33.6 & 61.9& 29.0 &56.1 & 61.7 & 81.0& 54.8 &76.3 & 44.9 & 70.5\\
\rowcolor{Gray} \oursm & \textbf{42.6} &\textbf{69.3} & \textbf{35.1} &\textbf{62.8} & \textbf{30.7} &\textbf{57.6} & \textbf{63.1} &\textbf{81.8} & \textbf{56.0} &\textbf{77.1} & \textbf{46.4} & \textbf{71.3} \\
\Xhline{2\arrayrulewidth}
\multicolumn{13}{c}{Text-Only Machine Translation} \\ \Xhline{2\arrayrulewidth} 
% MultiTask~\cite{elliott2017imagination} & 36.8 & 55.8 & -- & -- & -- & -- & -- & -- & -- & -- & -- & --\\
VMMT\textsubscript{F}~\cite{calixto2018latent} & 37.7 & 56.0 & 30.1 & 49.9 & 25.5 & 44.8 & -- & -- & -- & -- & -- & --\\
UVR-NMT~\cite{zhang2020neural} & 36.9 & -- & 28.6 & -- & -- & -- & 58.3 & -- & 48.7 & -- & -- & --\\
ImagiT~\cite{long2021generative} & 38.5 &55.7 & 32.1 & 52.4 & 28.7 &48.8 & 59.7 &74.0 & 52.4 &68.3 & 45.3 & 65.0\\
\ours & \textbf{41.9} & \textbf{68.8}  & \textbf{34.0} & \textbf{62.5} & \textbf{30.3} & \textbf{57.0} & \textbf{62.3} & \textbf{81.4} & \textbf{55.1} & \textbf{76.4} & \textbf{45.7} & \textbf{70.9}\\
\hline
\rowcolor{Gray} RMMT~\cite{wu2021good} & 41.4 & 68.0 & 32.9 & 61.7 & 30.0 & 56.3 & 62.1 & 81.3 & 54.4 & 76.1 & 44.5 & 70.2\\
\rowcolor{Gray} \ours & \textbf{42.7} & \textbf{69.3} & \textbf{35.1} & \textbf{62.8} & \textbf{30.7} & \textbf{57.5} & \textbf{63.1} & \textbf{81.8} & \textbf{56.0} & \textbf{77.1} & \textbf{46.5} & \textbf{71.4} \\
\Xhline{2\arrayrulewidth}
\end{tabular}}
\caption{\textbf{Comparison with state-of-the-art multimodal and text-only translation methods on Multi30K}. 
\ours hallucinates visual representations from text-only inputs, while \oursm uses ground-truth visual tokens at test time. 
Results in gray are computed with model averaging over $10$ latest checkpoints. 
\ours establishes new state-of-the-art for machine translation on Multi30K.
% outperforms all existing methods on three test sets.
% $^\dagger$ indicates model requires external visual data at test time.
% Our proposed approach \ours establishes new state-of-the-art for machine translation on Multi30K, by outperforming existing methods on all three test sets.
}
\label{tab:m30k_sota} \vspace{-3mm}
\end{table*}

\vspace{1mm}
\noindent\textbf{Baselines.} We compare 
% our approach 
with the following baselines. (1) text-only baseline that trains a transformer~\cite{vaswani2017attention} without any visual information, 
% for machine translation, 
(2) conventional MMT models (e.g., DCCN~\cite{lin2020dynamic}, GMNMT~\cite{yin2020novel}, and Gated Fusion~\cite{wu2021good}) that rely on sentence-image pairs for inference, (3) exiting methods where only text inputs are provided at test time for translation, including ImagiT~\cite{long2021generative}, UVR-NMT~\cite{zhang2020neural}, and RMMT~\cite{wu2021good}. 
We directly quote numbers reported in published papers when possible and use 
% author’s provided source 
publicly available codes for UVR-NMT and RMMT on both WIT and WMT datasets.    

% We compare \ours to baseline machine translation systems under two different setups. The first is unimodal inference, where only text inputs are provided at test time

\vspace{1mm}
\noindent\textbf{Evaluation Metrics.} We compute  BLEU~\cite{papineni2002bleu}
% \footnote{\url{https://github.com/moses-smt/mosesdecoder/blob/RELEASE-4.0/scripts/generic/multi-bleu.perl}} 
and METEOR~\cite{denkowski2014meteor}
% \footnote{\url{https://github.com/cmu-mtlab/meteor}} 
scores to measure the translation performance of different models. Unless otherwise noted, we select the checkpoint with lowest validation loss for inference and further average the last ten checkpoints as in~\cite{wu2021good,vaswani2017attention}, to compare with Gated Fusion/RMMT on Multi30K dataset.

\begin{table*}[ht]
\small
\centering

\resizebox{0.9\linewidth}{!}{
\begin{tabular}{@{}l|ccc|cc|cc|c@{}}
\Xhline{2\arrayrulewidth}
\multirow{2}{*}{\textbf{Method}} & \multicolumn{3}{c|}{\textbf{Well-Resourced}}    & \multicolumn{2}{c|}{\textbf{Non-English}}  & \multicolumn{2}{c|}{\textbf{Under-Resourced}} & \multirow{2}{*}{\textbf{Average}}    \\
\cline{2-8} 
 & \textbf{EN $\rightarrow$ DE}        & \textbf{EN $\rightarrow$ ES}     & \textbf{EN $\rightarrow$ FR}        & \textbf{DE $\rightarrow$ ES}    & \textbf{ES $\rightarrow$ FR}        & \textbf{EN $\rightarrow$ RO} & \textbf{EN $\rightarrow$ AF} &    \\ \Xhline{2\arrayrulewidth}
Text-Only   & 16.0 ± 0.5          & 24.8 ± 0.8          & 16.1 ± 1.2    & 10.7 ± 0.2  & 16.2 ± 0.3  & 11.5 ± 0.7 & 10.8 ± 0.6   & 15.1 ± 0.6 \\
% UVR-NMT~\cite{zhang2020neural}     & 16.2 ± 0.1          & 23.9 ± 1.5          & 16.3 ± 1.0          & 11.0 ± 0.1         &  16.3 ± 0.5         & 12.2 ± 0.2  & 9.3 ± 1.1         & 15.0 ± 0.6 \\ 
UVR-NMT~\cite{zhang2020neural}     & 16.9 ± 0.2          & 26.4 ± 0.4          & 17.7 ± 0.3          & 10.9 ± 0.9         &  16.4 ± 0.6         & 12.5 ± 0.5  & 11.6 ± 1.7         & 16.1 ± 0.7 \\ 
% RMMT~\cite{wu2021good}    & 15.7 ± 0.1          & 24.8 ± 0.4         & 15.9 ± 1.6         & 10.9 ± 0.3          & 15.9 ± 0.7         & 9.8 ± 1.5  & 9.5 ± 0.9         & 14.6 ± 0.8  \\
RMMT~\cite{wu2021good}    & 16.4 ± 0.3          & 24.8 ± 0.4         & 17.2 ± 1.6         & 11.0 ± 0.3          & 15.9 ± 0.7         & 9.9 ± 1.4  & 9.8 ± 1.0         & 15.0 ± 0.7  \\
% Gated Fusion~\cite{wu2021good}        &           &  22.5 ± 0.1        &    16.2 ± 0.9      & 10.5 ± 0.2          & 15.7 ± 0.4         & 9.9 ± 0.1  &          &  \\
\ours   & \textbf{17.5 ± 0.4}   & \textbf{27.5 ± 0.2}         & \textbf{18.8 ± 0.2}          & \textbf{11.3 ± 0.2}        & \textbf{16.6 ± 0.8}          & \textbf{14.4 ± 1.0} & \textbf{14.0 ± 0.5}          & \textbf{17.2 ± 0.4} \\
\oursm  & 17.4 ± 0.4  & \textbf{27.5 ± 0.2}  & \textbf{18.8 ± 0.2}          & \textbf{11.3 ± 0.2}         & \textbf{16.6 ± 0.8}          & \textbf{14.4 ± 1.0} & \textbf{14.0 ± 0.4}          & \textbf{17.2 ± 0.4} \\
\Xhline{2\arrayrulewidth}
\end{tabular}}
% \vspace{-1mm}
\caption{\textbf{BLEU score on WIT}. 
% \ours achieves an average $3.1$ score improvement over text-only baseline in under-resource setting including best average performance among all compared methods. 
Please refer to appendix for METEOR score comparisons.}
\label{tab:wit}
\end{table*}
\begin{table*}[ht]
\centering
\resizebox{\linewidth}{!}{
\begin{tabular}{l|c|cccc|cccc}
\Xhline{2\arrayrulewidth} 
\multirow{3}{*}{\textbf{Method}}                  & \multirow{3}{*}{\textbf{Visual   Data}} & \multicolumn{4}{c|}{\textbf{Well-Resourced}}                                              & \multicolumn{4}{c}{\textbf{Under-Resourced}}                                          \\ \cline{3-10} 
                                       &                                & \multicolumn{2}{c|}{\textbf{EN $\rightarrow$ DE}}                 & \multicolumn{2}{c|}{\textbf{EN $\rightarrow$ FR}} & \multicolumn{2}{c|}{\textbf{EN $\rightarrow$ DE}}                     & \multicolumn{2}{c}{\textbf{EN $\rightarrow$ FR}} \\ \cline{3-10} 
                                       &                                & \textbf{BLEU}         & \multicolumn{1}{c|}{\textbf{METEOR}} & \textbf{BLEU}            & \textbf{METEOR}   & \textbf{BLEU}         & \multicolumn{1}{c|}{\textbf{METEOR}}     & \textbf{BLEU}         & \textbf{METEOR}     \\ \Xhline{2\arrayrulewidth} 
Text-Only                              & --                             & 27.1 ± 0.2 & \multicolumn{1}{c|}{55.0 ± 0.1}      & 39.1 ± 0.2    & 64.4 ± 0.1        & 16.7 ± 0.2 & \multicolumn{1}{c|}{43.6 ± 0.2}          & 25.9 ± 0.1 & 52.3 ± 0.3         \\ \hline
UVR-NMT~\cite{zhang2020neural}& \multirow{2}{*}{Multi30K}      & 27.2 ± 0.2 (28.1)           & \multicolumn{1}{c|}{55.3 ± 0.1}      & 39.7 ± 0.2 (39.6)              & 64.9 ± 0.1        & 17.1 ± 0.1 & \multicolumn{1}{c|}{44.1 ± 0.3}          & 26.1 ± 0.3 & 52.8 ± 0.3          \\
RMMT~\cite{wu2021good}                                   &                                & 24.5 ± 0.2            & \multicolumn{1}{c|}{52.8 ± 0.1}      & 35.3 ± 0.0               & 61.2 ± 0.1        & 15.7 ± 0.2            & \multicolumn{1}{c|}{41.9 ± 0.4}          & 24.2 ± 0.3            & 50.7 ± 0.3          \\ \hline
\multirow{2}{*}{\ours}  & Multi30K                       & \textbf{28.0 ± 0.1}            & \multicolumn{1}{c|}{56.0 ± 0.1}      & \textbf{40.0 ± 0.1}               & \textbf{65.2 ± 0.1}        & 17.6 ± 0.1            & \multicolumn{1}{c|}{\textbf{44.8 ± 0.1}}          & \textbf{26.9 ± 0.2}            & 53.2 ± 0.2         \\
& WIT                            &  \textbf{28.0 ± 0.1}            & \multicolumn{1}{c|}{\textbf{56.1 ± 0.1}}      & 39.9 ± 0.1                & 65.1 ± 0.1        & \textbf{17.7 ± 0.2}  &  \multicolumn{1}{c|}{44.7 ± 0.1}          & 26.8 ± 0.0            & \textbf{53.3 ± 0.1}           \\
\hline
                                       
\multirow{2}{*}{\oursm}  & Multi30K                       & \textbf{28.0 ± 0.0}            & \multicolumn{1}{c|}{56.0 ± 0.1}      & 39.9 ± 0.1               &   65.0 ± 0.1     & \textbf{17.7 ± 0.1}            & \multicolumn{1}{c|}{\textbf{44.8 ± 0.2}}          & \textbf{26.9 ± 0.2}           & \textbf{53.3 ± 0.3}          \\
& WIT                            & 27.9 ± 0.1           & \multicolumn{1}{c|}{56.0 ± 0.2}      & 39.8 ± 0.2               &  65.0 ± 0.1       & \textbf{17.7 ± 0.2}           & \multicolumn{1}{c|}{\textbf{44.8 ± 0.1}}          & 26.8 ± 0.1            & \textbf{53.3 ± 0.1}          \\
% \multirow{3}{*}{\oursr} & Multi30K                       & ±            & \multicolumn{1}{c|}{±}      & ±               & ±        & 17.62 ± 0.07 & \multicolumn{1}{c|}{44.8 ± 0.1} & 26.88 ± 0.21 & 53.2 ± 0.2 \\
%                                       & WIT                            & ±            & \multicolumn{1}{c|}{±}      & ±               & ±        & ±            & \multicolumn{1}{c|}{±}          & ±            & ±          \\
%                                       & M30K + WIT                     & ±            & \multicolumn{1}{c|}{±}      & ±               & ±        & ±            & \multicolumn{1}{c|}{±}          & ±            & ±          \\ \hline
% \multirow{2}{*}{\oursm} & Multi30K                       & ±            & \multicolumn{1}{c|}{±}      & ±               & ±        & 17.66 ± 0.09 & \multicolumn{1}{c|}{44.8 ± 0.2} & 26.91 ± 0.22 & 53.3 ± 0.3 \\
%                                       & WIT                            & ±            & \multicolumn{1}{c|}{±}      & ±               & ±        & ±            & \multicolumn{1}{c|}{±}          & ±            & ±          \\
%                                       & M30K + WIT                     & ±            & \multicolumn{1}{c|}{±}      & ±               & ±        & ±            & \multicolumn{1}{c|}{±}          & ±            & ±          \\ 
\Xhline{2\arrayrulewidth} 
\end{tabular}}

\caption{\textbf{Results on WMT2014}. 
% Our proposed approach, \ours outperforms all the compared methods in both well-resourced and under-resourced settings. 
UVR-NMT results in brackets are reported by the original paper.}
\label{tab:wmt} \vspace{-2mm}
\end{table*}

% \vspace{-1mm}
% \vspace{-0.5mm}
\subsection{Results on Multi30K}
% \vspace{-1mm}
Table~\ref{tab:m30k_ours} shows the results on Multi30K. 
% of our approach and text-only baselines . 
Transformer-Tiny ($\sim20$ times smaller than Transformer-Base) obtains the best performance in text-only translation, which is consistent with the recent findings in~\cite{wu2021good}. \ours (denoted by \texttt{V} in Table~\ref{tab:m30k_ours}) significantly outperforms the text-only baselines on all three test sets, which demonstrates the effectiveness of visual hallucination for text-only NMT. Using Transformer-Tiny as the backbone, \ours obtains an average $35.4$ BLEU in EN$\rightarrow$DE and $54.4$ BLEU in EN$\rightarrow$FR, which is about $2.1$ and $1.4$ BLEU improvements over the text-only baseline. 
Moreover, \ours has very similar performance 
with either hallucinated (\texttt{V}) or ground-truth representation (\texttt{VM}), showing strong ability to generate visual representations that are semantically consistent with the ground-truth.

Table~\ref{tab:m30k_sota} shows that 
% our proposed approach 
\ours outperforms all compared methods, achieving best BLEU and METEOR scores under both mulitmodal and text-only translation settings. While comparing to ImagiT~\cite{long2021generative}, that generates continuous hallucinations via adversarial learning, \ours obtains $2.3$ and $1.9$ BLEU improvements on the EN$\rightarrow$DE and EN$\rightarrow$FR tasks respectively, showing the effectiveness of discrete visual representations. Similarly, \ours significantly outperforms UVR-NMT~\cite{zhang2020neural} in both tasks, without relying on additional image retrieval at test time. In summary, these consistent improvements clearly show that \ours can effectively leverage visual semantics available at training time to greatly improve text-only test time translation.

\begin{figure}
    \centering
    \includegraphics[width=0.45\linewidth]{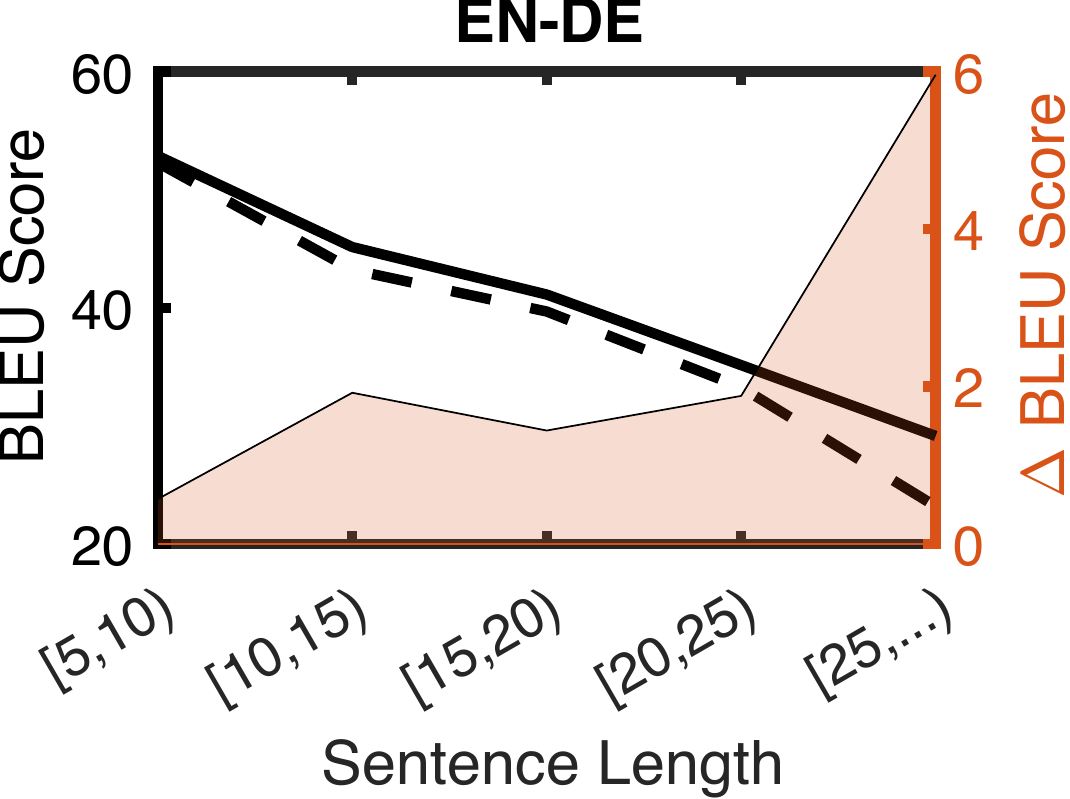} \quad
    \includegraphics[width=0.45\linewidth]{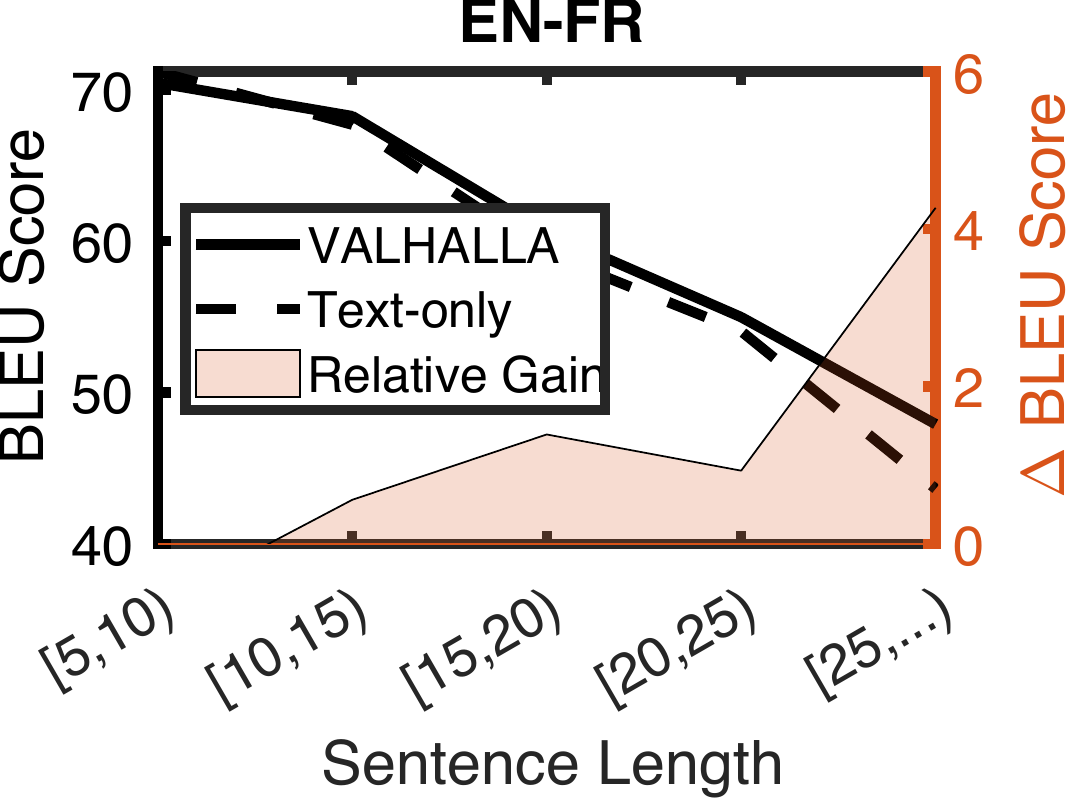}
    \caption{BLEU scores on different groups divided according to source sentence lengths on Multi30K Test2016 split.}
    \label{fig:input_length}
\end{figure}

We further divide the Test2016 set into different groups based on  lengths of source sentences and compare performance with a text-only baseline in each group, as shown in Figure~\ref{fig:input_length}. \ours consistently achieves the best performance in all groups, which once again confirms the effectiveness and generality of our approach. We further observe that the improvements are particularly pronounced for long sentences on both EN$\rightarrow$DE and EN$\rightarrow$FR tasks.
% potentially because long sentences often contain more ambiguous words requiring visual context for better translation. 

\vspace{-1mm}
\subsection{Results on WIT} 
% \vspace{-0.5mm}

Table~\ref{tab:wit} shows that on WIT, \ours again outperforms existing methods, improving text-only baseline performance from $15.1$ to $17.2$ BLEU, (see Table~\ref{tab:wit_meteor} for METEOR scores). In particular, our approach achieves a substantial improvement over text-only baseline in under-resourced settings ($2.9$ on EN$\rightarrow$RO and $3.2$ on EN$\rightarrow$AF). This shows that \ours is more robust to conditions where the training corpora is small, revealing an important advantage of grounding information provided by visual hallucination for machine translation. Interestingly, while our approach is overall effective in translation between non-English languages, the improvement over text-only baseline is marginal. This is potentially due to an English-centric bias in the image-text pairs of original WIT dataset, which might mean that visual modality fails to provide much additional information for translation between non-English languages.

\begin{figure}
    \centering
    \includegraphics[width=0.48\linewidth]{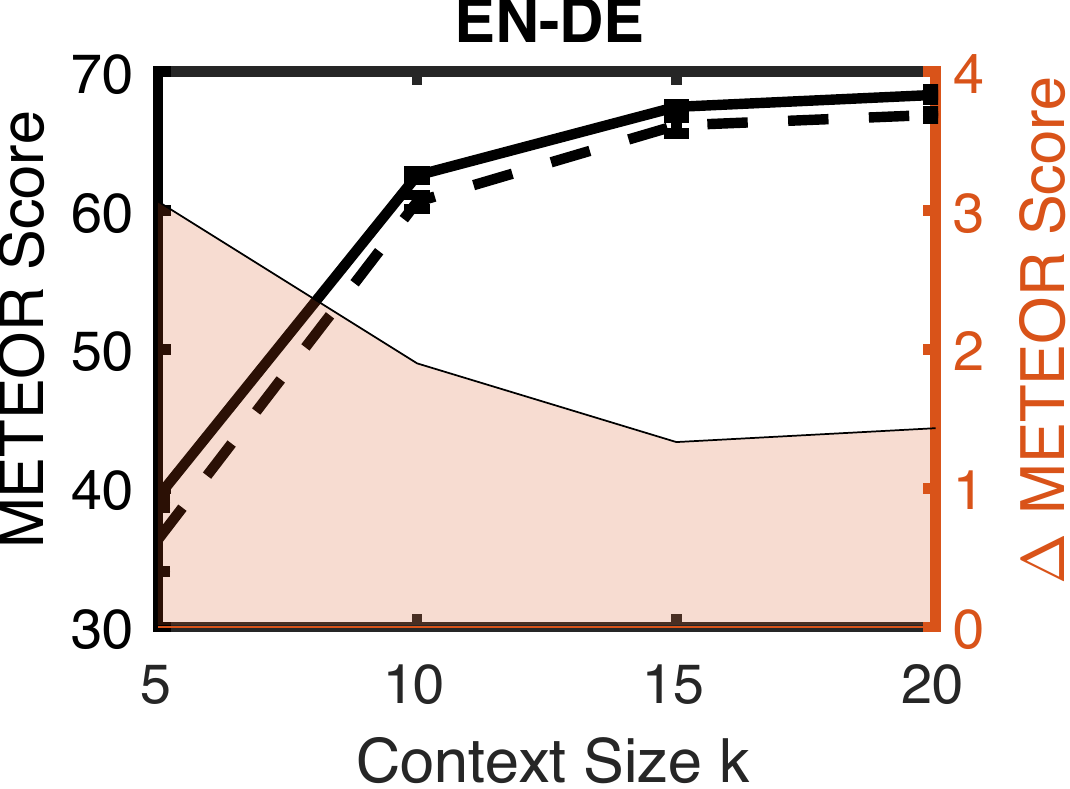} \hfill
    \includegraphics[width=0.48\linewidth]{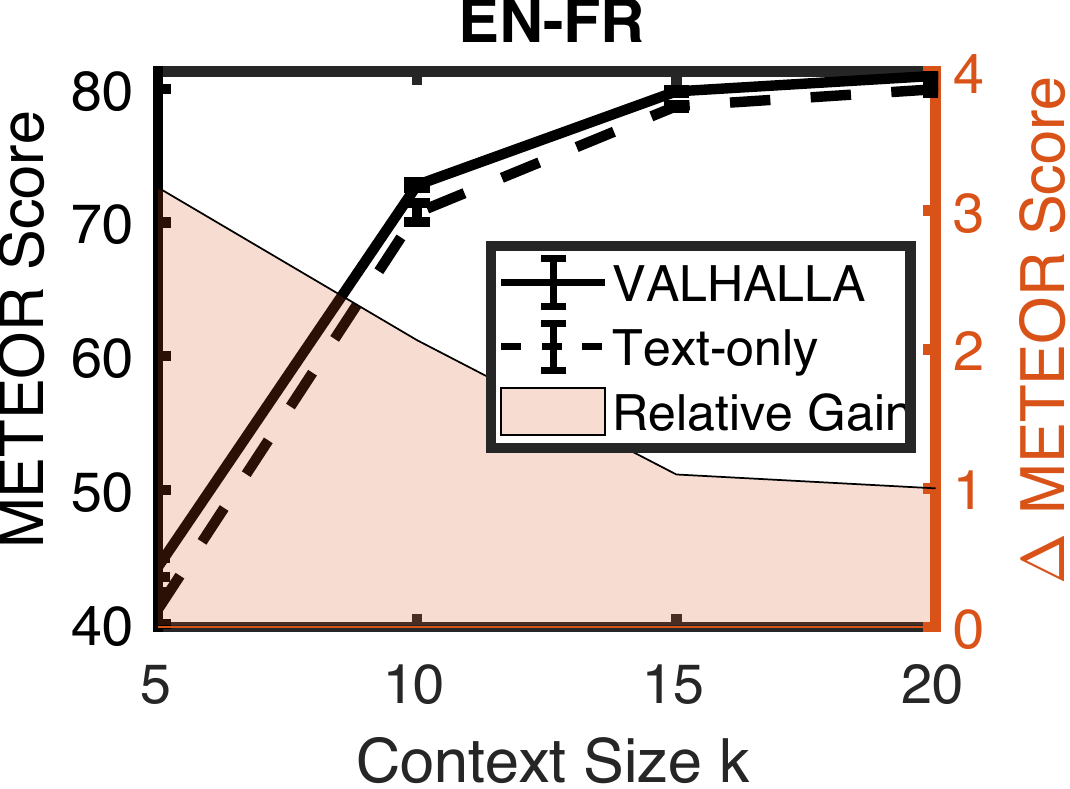}
    \caption{\textbf{Evaluation with Progressive Masking.} 
    All results use METEOR scores on Multi30K Test2016 split.
    % We plot the METEOR scores of \ours and text-only models evaluated on Test2016 split, as well as the relative improvements over baseline.
    }
    \label{fig:progressive_mask}
\end{figure}

\vspace{-1mm}
\subsection{Results on WMT}
% \vspace{-0.5mm}

Table~\ref{tab:wmt} shows the results on WMT.
\ours benefits from visual hallucination and outperforms all the compared methods in both well- and under-resourced settings.
The improvements over text-only baseline are more significant in under-resourced scenarios, which is of significant practical value.
We find that use of larger datasets, e.g., WIT instead of Multi30K for retrieving images at training time does not lead to substantial gain in performance, which is consistent with the previous findings~\cite{zhang2020neural}.
Overall, the results on WMT show that our approach can be integrated into large-scale text-only translation datasets representing a wide variety of abstract concepts and real world entities (i.e., not specifically designed for multimodal machine translation).

% \vspace{-1mm}
\subsection{Translation Under Limited Textual Context}
% \vspace{-1mm}
% Following the experimental setup of~\cite{caglayan2019probing}, 
We further study the robustness of \ours framework for machine translation under limited textual context by degrading the input language modality during training and inference in two ways~\cite{caglayan2019probing}:
% simulating low-resource conditions where sentences may miss crucial information~\cite{caglayan2019probing}.
% We consider two types of degradations to the input language modality by selectively masking input tokens during training and inference:
(1) Progressive masking that replaces all but the first $k$ words of source sentences with a special token \texttt{<v>}, (2) Visual entity masking that randomly replaces visually grounded phrases (annotation from~\cite{plummer2015flickr30k}) with probability $p$ in the source sentence with \texttt{<v>}.

\begin{figure}
    \centering
    \includegraphics[width=0.48\linewidth]{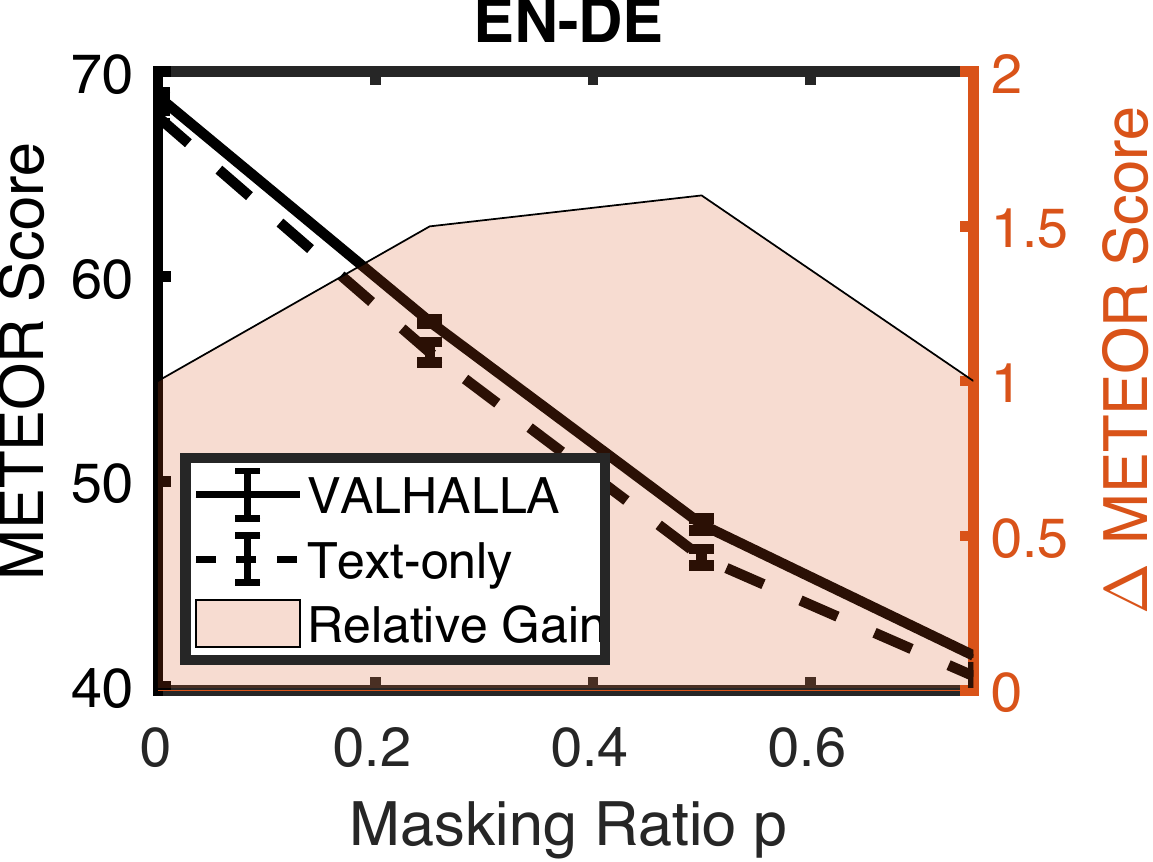} \hfill
    \includegraphics[width=0.48\linewidth]{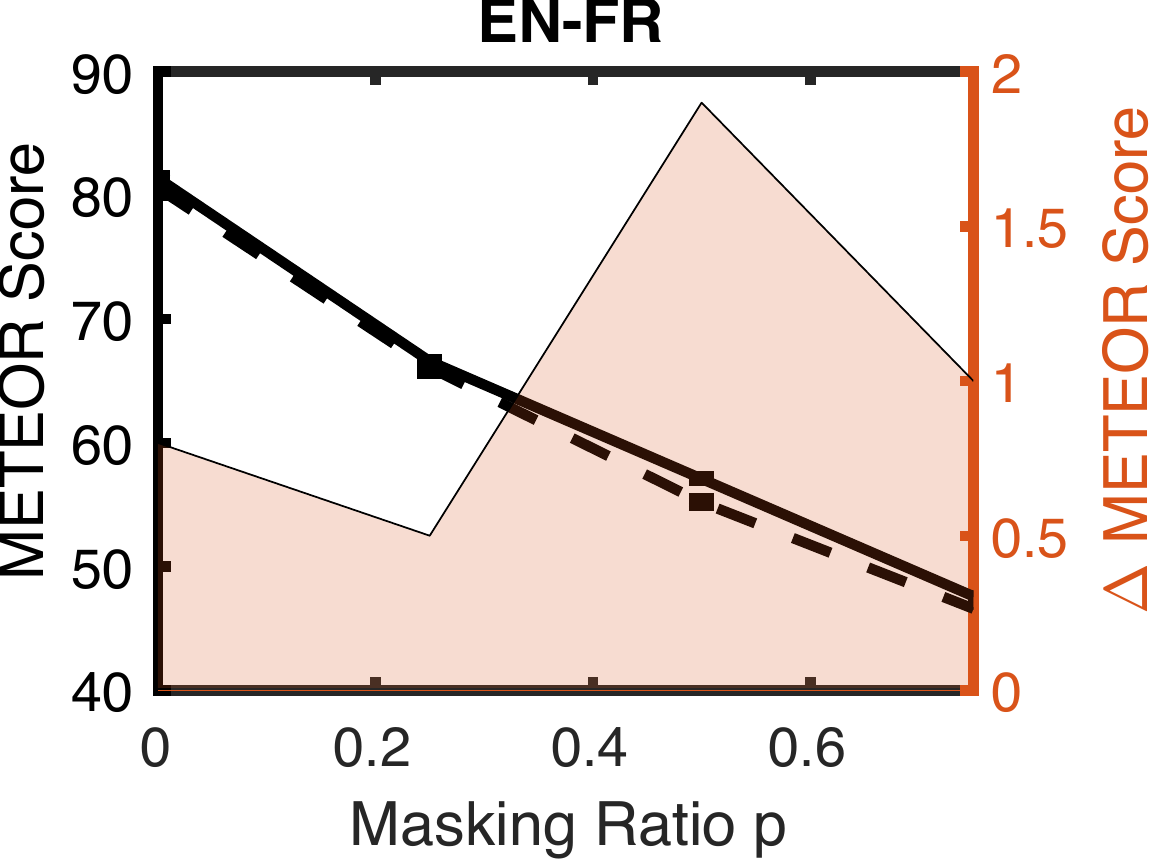}
    \caption{\textbf{Evaluation with Entity Masking.} 
    All results use METEOR scores on Multi30K Test2016 split. 
    % We plot the METEOR scores of \ours and text-only models evaluated on Test2016 split, as well as the relative improvements over baseline.
    }
    \label{fig:entity_mask}
    \vspace{-2mm}
\end{figure}

\vspace{0.5mm}
\noindent\textbf{Progressive Masking.} Figure~\ref{fig:progressive_mask} compares METEOR score of text-only baseline and \ours as a function of context length $k$. On both EN$\rightarrow$DE and EN$\rightarrow$FR tasks, \ours consistently outperforms the baseline under all settings. The gap between both methods widens as context size is reduced, with \ours performing $\sim 3$ METEOR points better. This suggests that visual hallucination is even more effective for translating ambiguous sentences out of context.

\begin{table*}
    \centering
    \begin{subtable}[b]{0.44\linewidth}
        \centering
        \newcommand{\xmark}{\ding{55}}%

\resizebox{\linewidth}{!}{
\begin{tabular}{@{}lccccc@{}}
\toprule
\textbf{Backbone}          & \textbf{\begin{tabular}[c]{@{}c@{}}Discrete\\ Embedding\end{tabular}} & \textbf{\begin{tabular}[c]{@{}c@{}}External\\ Pretraining\end{tabular}} & \textbf{Aggregation} & \textbf{EN-DE} & \textbf{EN-FR} \\ \midrule
CLIP RN-50                 & \xmark                                                                & CLIP                                                                    & Gating               & 38.0           & 58.8           \\ \midrule
\multirow{2}{*}{ResNet-50} & \multirow{2}{*}{\xmark}                                               & \multirow{2}{*}{ImageNet}                                               & Gating               & 38.8           & 59.1           \\
                           &                                                                       &                                                                         & Concatenation        & 38.3           & 60.0           \\ \midrule
VQGAN VAE                  & \checkmark                                                            & None                                                                    & Concatenation        & \textbf{39.6}  & \textbf{60.5}  \\ \bottomrule
\end{tabular}
}
\caption{Discrete and continuous visual encoder backbones, evaluated with Transformer-Small on Multi30K Test2016 split.}
\label{tab:ablations_vis_embed}
    \end{subtable} \hfill
    \begin{subtable}[b]{0.28\linewidth}
        \centering
        \resizebox{\linewidth}{!}{
% \begin{tabular}{@{}ccccc@{}}
% \toprule
% \textbf{\begin{tabular}[c]{@{}c@{}}Input  \\ Size\end{tabular}} &
%   \textbf{\begin{tabular}[c]{@{}c@{}}Encoder\\ Layers\end{tabular}} &
%   \textbf{\begin{tabular}[c]{@{}c@{}}Visual Token\\ Length\end{tabular}} &
%   \textbf{EN-DE} &
%   \textbf{EN-FR} \\ \midrule
% \multirow{3}{*}{$128   \times 128$} & 4 & $16^2   = 256$   & ± & ± \\
%                                     & 5 & $8^2   = 64$     & ± & ± \\
%                                     & 6 & $4^2   = 16$     & ± & ± \\ \midrule
% \multirow{3}{*}{$256   \times 256$} & 5 & $16^2   =   256$ & ± & ± \\
%                                     & 6 & $8^2   =   64$   & ± & ± \\
%                                     & 7 & $4^2   =   16$   & ± & ± \\ \bottomrule
% \end{tabular}
\renewcommand{\arraystretch}{1.4}
\begin{tabular}[b]{@{}cccc@{}}
\toprule
  \textbf{\renewcommand{\arraystretch}{1}\begin{tabular}[c]{@{}c@{}}Encoder\\ Layers\end{tabular}} &
  \textbf{\renewcommand{\arraystretch}{1}\begin{tabular}[c]{@{}c@{}}Visual Token\\ Length\end{tabular}} &
  \textbf{EN-DE} &
  \textbf{EN-FR} \\ \midrule
4 & $16^2   = 256$   & 13.5 ± 7.2 & 54.3 ± 0.4 \\
5 & $8^2   = 64$     & 36.3 ± 0.2 & 60.3 ± 0.2 \\
6 & $4^2   = 16$     & \textbf{39.6 ± 0.3} & \textbf{60.5 ± 0.1} \\ \bottomrule
\end{tabular}
}
\caption{Visual encoder depths, evaluated with Transformer-Small on Multi30K Test2016.
}
\label{tab:ablations_encoder_size}
    \end{subtable} \hfill
    % \\
    % \begin{subtable}{0.45\linewidth}
    %     \centering
    %     \input{tables/ablations/clip}
    % \end{subtable} \hfill
    \begin{subtable}[b]{0.25\linewidth}
        \centering
        \newcommand{\xmark}{\ding{55}}%

\resizebox{\linewidth}{!}{
\begin{tabular}[b]{@{}cccc@{}}
\toprule
\textbf{\begin{tabular}[c]{@{}c@{}}Visual\\ Data\end{tabular}} & \textbf{\begin{tabular}[c]{@{}c@{}}Image\\ Retrieval\end{tabular}} & \textbf{EN-DE} & \textbf{EN-FR} \\ \midrule
\multirow{2}{*}{Multi30K} &  \xmark          & 16.5 ± 0.3 & 26.2 ± 0.1 \\
                          & \checkmark & \textbf{17.6 ± 0.1} & \textbf{26.9 ± 0.2} \\ \midrule
\multirow{2}{*}{WIT}      &    \xmark         & 16.6 ± 0.2 & 26.1 ± 0.3 \\
                          & \checkmark & \textbf{17.7 ± 0.2} & \textbf{26.8 ± 0.0} \\ \bottomrule
% \multirow{2}{*}{Multi30K} &            & 16.52 & 26.15 \\
%                           & \checkmark & 17.62 & 26.88 \\ \midrule
% \multirow{2}{*}{WIT}      &            & 16.61 & 26.07 \\
%                           & \checkmark & 17.68 & 26.82 \\ \bottomrule
\end{tabular}
}
\caption{Training on WMT under-resourced tasks \emph{without} image retrieval.}
\label{tab:ablations_retrieval}
    \end{subtable}
    % \begin{subtable}{0.85\linewidth}
    %     \centering
    %     \input{tables/ablations/external_data}
    % \end{subtable}
    \caption{\textbf{Ablation Studies.} All results use BLEU scores.}
    \label{tab:ablations} \vspace{-3mm}
\end{table*}

\begin{figure}
    \centering
    \includegraphics[width=\linewidth]{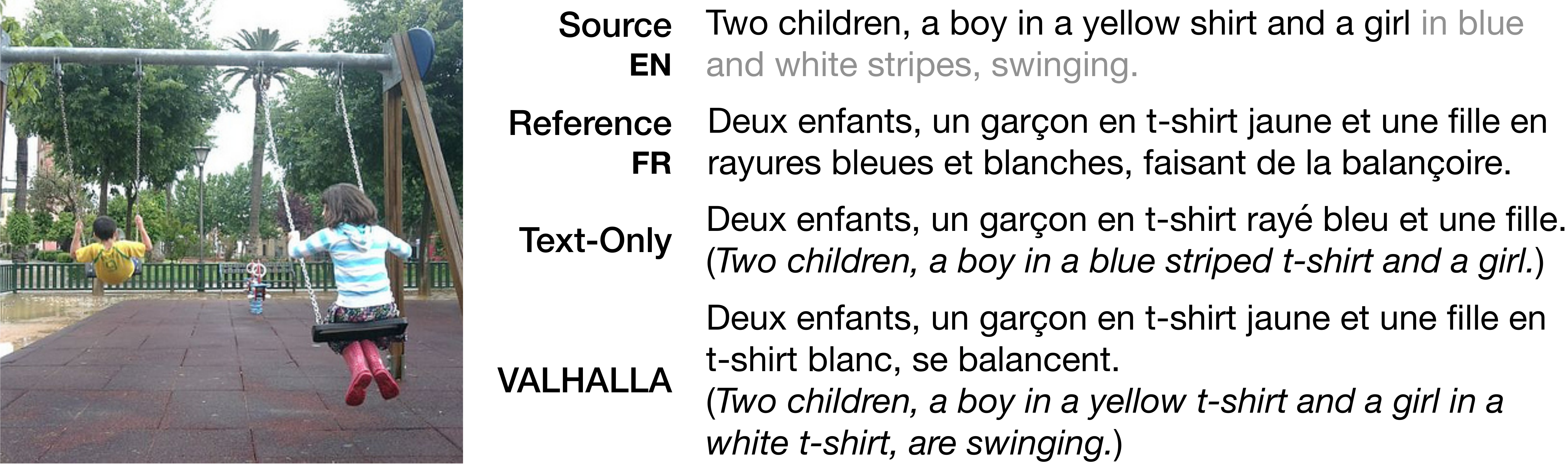} 
    % \\[2mm]
% \includegraphics[width=\linewidth]{figures/images/qualitative2-cropped.pdf}
    \caption{\textbf{Qualitative Result with Progressive Masking.} Phrases in gray in the source sentence are masked with \texttt{<v>} at model input.
    % See supplemental for more examples.
    }
    \label{fig:qualitative}
\end{figure}

\vspace{0.5mm}
\noindent\textbf{Visual Entity Masking.} 
Figure~\ref{fig:entity_mask} compares \ours with text-only baseline when visual entities from the input source sentences are masked with probability $p$. Again, \ours beats the text-only baseline in all test cases, with greatest improvements observed at $p = 0.5$. We attribute this to the effect of hallucination transformer inherently modeling co-occurence between visual entities (e.g. human and objects) in the scene. This advantage reduces as masking ratio is further increased to $0.75$, likely due to inability of visual hallucination to generate plausible predictions when majority of visual concepts are missing from input sentences.

\vspace{0.5mm}
\noindent\textbf{Qualitative Examples.} 
% Figure~\ref{fig:qualitative} shows sample translation outputs from \ours and baseline models under limited textual context. \todo{Discussion}
Figure~\ref{fig:qualitative} shows sample translation outputs from \ours and text-only baseline under progressive masking, where \ours successfully predicts masked phrase ``swinging'' through visual hallucination. 
% More qualitative examples 
% of progressive and visual entity masking 
% are included in the supplemental.

% \vspace{-1mm}
\subsection{Ablation Analysis}
\vspace{-1mm}

% We present the following ablation analysis on different components of our framework, including visual encoder designs, hallucination strategies, and optimization details.

\vspace{0.5mm}
\noindent\textbf{Discrete vs. Continuous Representations.} 
% We first examine the efficacy of discrete visual encoder for multimodal translation, comparing it with conventional continuous encoders (\eg ResNet~\cite{he2016deep}), as well as alternative discrete VAE models. Table~\ref{tab:ablations_vis_embed} compares translation performance of the VQGAN backbone to pretrained ResNet-50 \todo{and the discrete VAE model of DALL-E~\cite{ramesh2021zero}} When using the ResNet backbone, we use the \texttt{conv5} feature maps before global pooling, flattened into a sequence of $7 \times 7$ continuous feature vectors. \todo{Discussion}
% We examine the efficacy of discrete visual representations for multimodal translation, comparing it with conventional continuous representations (\eg, ResNet~\cite{he2016deep}) under different aggregation schemes. 
Table~\ref{tab:ablations_vis_embed} compares performance of discrete VQGAN VAE 
% to that of conventional 
with continuous representations, e.g., ResNet-50~\cite{he2016deep} pretrained on ImageNet~\cite{krizhevsky2012imagenet} 
% for ImageNet classification~\cite{krizhevsky2012imagenet} 
or CLIP~\cite{radford2021learning}.
% image-text correspondence~\cite{radford2021learning}. 
For ResNet, we consider two strategies to fuse multimodal inputs: \emph{Gating}~\cite{yin2020novel,zhang2020neural,wu2021good} learns a 
% cross-modal 
gating layer between text embeddings and pooled ResNet features; \emph{Concatenation} flattens the feature map at \texttt{conv5} block before global pooling, projects to the dimension of text embeddings and concatenates them, similar to the strategy used by \ours with discrete tokens. 
% As seen from 
Table~\ref{tab:ablations_vis_embed} shows that ImageNet pretraining remains an effective way to extract continuous visual features, outperforming CLIP pretraining under the gating 
% aggregation 
strategy. While aggregation by concatenation has no clear benefit over gating for continuous representations, it produces the strongest results for a discrete visual encoder. Importantly, since this strategy avoids pretraining encoders on large external datasets, it is potentially generalizable to a wider range of applications.

% \vspace{1mm}
% \noindent\textbf{Vision-language aggregation.}
% Table~\ref{tab:ablations_vislang_agg}: Attention, Gated Fusion, Concatenation. \yl{Move to supplemental? We already compared to gated fusion in main results}

\vspace{1mm}
\noindent\textbf{Visual Encoder Design.} 
% We inspect the effect of visual encoder design on the translation scores by varying the depth of encoder, resulting in different lengths of encoded visual tokens. 
Table~\ref{tab:ablations_encoder_size} shows the effect of varying depth of visual encoder, resulting in different lengths of encoded visual tokens. A smaller visual sequence is beneficial for multimodal modeling, as too many visual tokens may prevent the translation transformer from attending to the relevant text sequence and overfit to image inputs instead.

\begin{figure}
    \centering
     \includegraphics[height=0.33\linewidth]{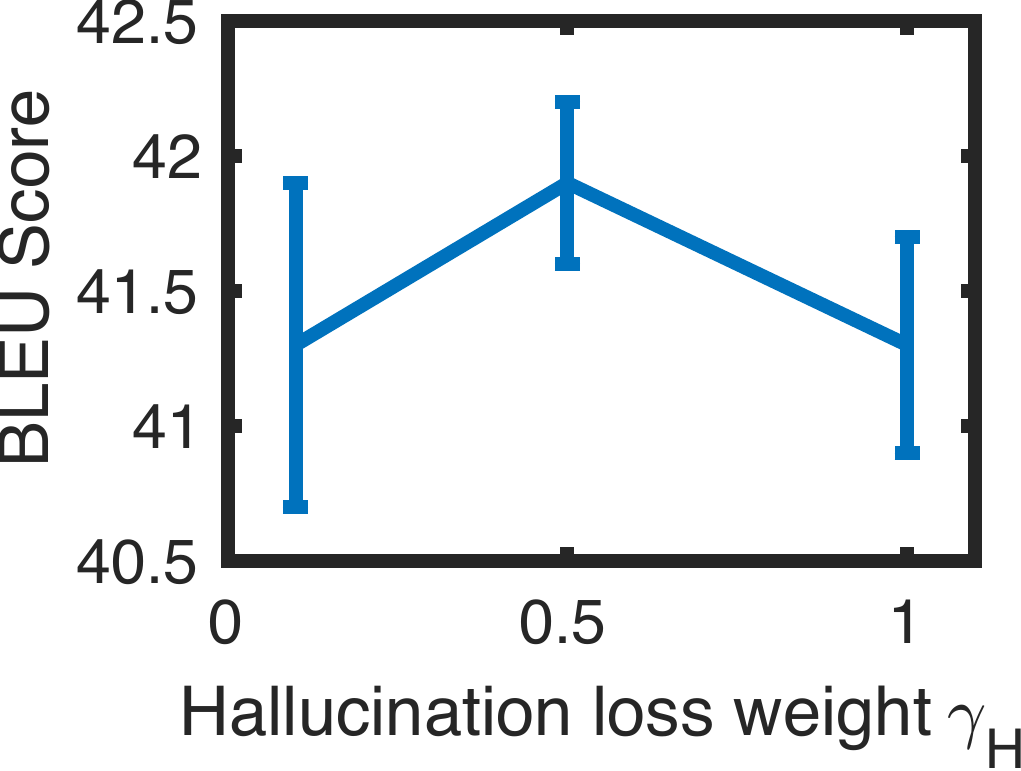} \qquad
    \includegraphics[height=0.33\linewidth]{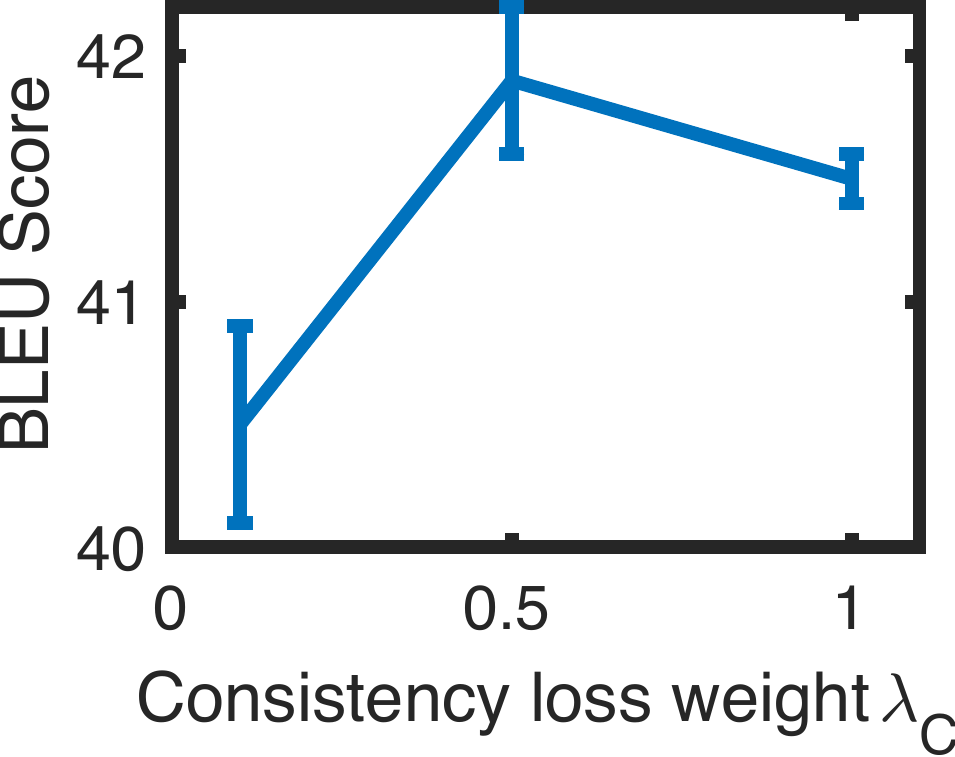}
    \caption{Influence of loss weights $\gamma_H$ and $\lambda_C$ of \eqref{eqn:valhalla_loss} on translation performance, measured on Multi30K EN$\rightarrow$DE task.}
    \label{fig:ablations_loss_weight}
\end{figure}

\vspace{1mm}
\noindent\textbf{Joint Optimization.}
Compared to the jointly trained \ours model, using a pretrained visual hallucination module off-the-shelf yields worse results (EN$\rightarrow$DE BLEU: $39.0$ vs $39.6$ and $31.0$ vs $31.7$ with Transformer-Small model on Multi30K Test2016 and Test2017 respectively). This validates the necessity of jointly fine-tuning the hallucination transformer $\mathbf{f_H}$ and translation transformer $\mathbf{f_T}$ with the Gumbel-softmax sampling strategy outlined in \eqref{eqn:gumbel}. 

\vspace{1mm}
\noindent\textbf{Randomized Visual Tokens.}
On Multi30K'16 EN$\rightarrow$FR, replacing inputs to MMT transformer $\bf f_T$ with \emph{random} visual tokens reduced BLEU score to $61.2$ from $62.3$. 
% within the error margin of text-only baseline (60.9). 
We observe a similar drop ($\sim$1--2 BLEU) in performance while using random visual tokens on other tasks as well, 
% when hallucinated features are replaced with random ones, 
which suggests that hallucinated visual tokens are indeed of crucial significance.

% We conduct a  Multi30K EN$\rightarrow$FR, replacing inputs to MMT transformer $\bf f_T$ with \emph{random} visual tokens reduced BLEU score to $61.2$ from $62.3$. \todo{Results on other tasks} The consistent drop in performance by using randomized tokens suggests that visual hallucination under limited textual context is indeed of crucial significance.

\vspace{1mm}
\noindent\textbf{Loss Hyperparameters.}
% Figure~\ref{fig:ablations_loss_weight} shows the sensitivity of performance to the weights of training losses introduced in \eqref{eqn:halluc_loss} and \eqref{eqn:halluc_consistency_loss}. 
Figure~\ref{fig:ablations_loss_weight} shows that 
% the model behavior 
\ours is robust to the choice of hallucination weight $\gamma_H$ but more sensitive to the consistency hyperparameter $\lambda_C$
% consistency hyperparameter $\lambda_C$ plays a more important role, where 
($1.4$ BLEU improvement when increasing $\lambda_C$ from $0.1$ to $0.5$). This shows that it is crucial to enforce consistency between translation \emph{outputs} based on ground-truth and hallucinated features \eqref{eqn:halluc_consistency_loss}, in addition to consistency \eqref{eqn:halluc_loss} in \emph{visual} latent space.

% \paragraph{Hallucinating multiple images.} \todo{Move to supplemental?}

% \paragraph{Random visual tokens.} \todo{Move to supplemental?}

% \vspace{1mm}
% \noindent\textbf{CLIP Hallucination.} A naive method to predict visual features from input sentence is to utilize a CLIP model~\cite{radford2021learning}, learned with a contrastive loss that aligns the embedding space of text and image. We trained a multimodal translation model with gating strategy on top of image features extracted from the pretrained CLIP visual encoder, and replace this with text embeddings from CLIP language encoder to realize text-only inference. 
% % Table~\ref{tab:ablations_clip} shows the result of this CLIP-based hallucination model\dots \todo{Discussion}
% The resulting model achieves a performance of \todo{Results: Hallucination, multimodal}

\vspace{1mm}
\noindent\textbf{Image Retrieval.} We study the importance of image retrieval in training with the text-only corpora of WMT. Table~\ref{tab:ablations_retrieval} shows that the performance of \ours trained with translation loss $\ell_T(\mathbf{f_T}; \hat{z})$ alone (i.e., directly using the hallucination transformer trained on Multi30K or WIT without retrieved images $v$) is considerably worse. This shows that the retrieved real images serve as important regularizer for the hallucination and translation transformers. 

% \vspace{1mm}
% \noindent\textbf{Training with External Data.} 
% % We also explore the possibility to further improve translation performance of \ours models by pretraining on larger datasets. To test this, 
% Experiments on WIT with a universal visual encoder (VQGAN VAE) trained on the union of images from all seven tasks increased performance on $4$ out of $7$ language pairs. However, the average improvement in BLEU score
% % the average improvement over all tasks is only 
% was marginal ($17.2$ vs $17.3$). This validates the sample efficiency of the VQGAN VAE, especially on the small training sets of low-resource settings. 
% with results shown in Table~\ref{tab:ablations_external} shows that 
% While training visual encoder with external data increases performance on 4 out of 7 language pairs, the average improvement over all tasks is only improved marginally by 0.1 BLEU score ($17.2$ vs $17.3$). This test validates the sample efficiency of the VQGAN VAE encoder, especially on small training sets under low-resource settings.

% More qualitative examples and implementation details
% are included in the supplemental.
% \vspace{-2mm}

%%%%%%%%% Conclusion %%%%%%%%%
\vspace{-0.5mm}
\section{Conclusion}
\label{sec:conclusion}
% \vspace{-0.5mm}

% In this paper, 
We present a new framework for improved machine translation by leveraging visual hallucination at test time, as opposed to existing MMT approaches based on sentence-image pairs. 
% In particular, 
We utilize an autoregressive hallucination transformer to generate discrete visual representations from the input text and train it jointly with a multimodal translation transformer. We demonstrate effectiveness of our 
% proposed 
approach on three 
% standard 
datasets, outperforming several competing methods.

% \vspace{1mm}
% \noindent\textbf{Limitations.} 
% Effectiveness of our approach depends on availability of good quality images to train the visual hallucination transformer, which is often difficult to collect especially for languages beyond English. Another potential limitation is training complexity which we believe could be greatly improved if we pre-extract VQGAN-VAE tokens, like existing methods did with ResNet-based visual encoders. 

% \vspace{0.5mm}
% \noindent\textbf{Broader Impact.}  
% Our research can have a positive impact on many real-world applications of neural machine translation involving a broad range of languages. It helps  performance in both well- and under-resourced scenarios which is of great practical importance. 
% Our research can have a positive impact in improving the performance of MT systems in under-resourced scenarios which is of great practical importance. Negative impacts of our research are difficult to predict, however, it shares many of the pitfalls associated with standard MT models such as dataset/social bias and susceptibility to adversarial attacks. While we believe that these issues should be mitigated, they are beyond the scope of this paper. \todo{Move to supplemental}

\vspace{1mm}
\noindent\textbf{Acknowledgements.} 
% This work was funded in part by NSF awards IIS-1924937, IIS-2041009, and a gift from Amazon.
This work was funded in part by NSF awards IIS-1924937 and IIS-2041009.
% \todo{Funding disclosure} 

\clearpage
{\small
\bibliographystyle{ieee_fullname}
\bibliography{ref,ref_supp}

%%%%%%%%% Supplementary %%%%%%%%%
\appendix
\section{Dataset Details}

We evaluate the performance of our proposed approach (\ours) using three machine translation datasets, namely Multi30K~\cite{elliott2016multi30k}, Wikipedia Image Text (WIT)~\cite{srinivasan2021wit} and WMT2014~\cite{bojar2014findings}.
These datasets present a diversity of challenges in machine translation: Multi30K requires models to learn to aggregate vision-language information from a relatively small number of training samples, while WIT and WMT contains translation tasks with different data scales. WMT additionally focuses on translating news articles, which may not be as readily grounded through visual data (compared to Multi30K and WIT), and thus presents an especially challenging test bed for MMT systems. 
Below we provide more details on each of the dataset.

\begin{table*}[ht]
    \centering
\resizebox{0.95\linewidth}{!}{
\begin{tabular}{l|cc|ccccccc|cccc}
\Xhline{2\arrayrulewidth}
\textbf{Dataset}          & \multicolumn{2}{c|}{Multi30K~\cite{elliott2016multi30k}}  & \multicolumn{7}{c|}{\multirow{2}{*}{WIT~\cite{srinivasan2021wit}}}  & \multicolumn{4}{c}{WMT2014~\cite{bojar2014findings}}       \\ 
\cline{1-3} \cline{11-14} 
\textbf{Visual Data}      & \multicolumn{2}{c|}{Flickr30K~\cite{young2014image}} & \multicolumn{7}{c|}{}                                                           & \multicolumn{4}{c}{Flickr30K~\cite{young2014image} / WIT~\cite{srinivasan2021wit}}                          \\ \Xhline{2\arrayrulewidth}
\textbf{Source Language}  & EN            & EN            & EN   & EN   & \multicolumn{1}{c|}{EN}   & DE   & \multicolumn{1}{c|}{ES}   & EN         & EN         & EN   & \multicolumn{1}{c|}{EN}  & EN            & EN            \\
\textbf{Target Language}  & DE             & FR            & DE   & ES   & \multicolumn{1}{c|}{FR}   & ES   & \multicolumn{1}{c|}{FR}   & RO         & AF         & DE   & \multicolumn{1}{c|}{FR}  & DE            & FR            \\ \hline
\textbf{\# Train Samples} & 29k            & 29k           & 329k & 287k & \multicolumn{1}{c|}{234k} & 133k & \multicolumn{1}{c|}{122k} & 40k        & 18k        & 3.9m & \multicolumn{1}{c|}{36m} & 100k          & 100k          \\
\textbf{\# Validation Samples} & 1k            & 1k           & 15k & 15k & \multicolumn{1}{c|}{15k} & 10k & \multicolumn{1}{c|}{10k} & 5k        & 5k        & 39k & \multicolumn{1}{c|}{27k} & 39k  & 27k          \\
\textbf{\# Test Samples} & 2.5k            &    2.5k        & 3k & 3k & \multicolumn{1}{c|}{3k} & 2k & \multicolumn{1}{c|}{2k} & 1k        & 1k        & 3k & \multicolumn{1}{c|}{3k} & 3k  &    3k       \\
\hline
\textbf{BPE Vocabulary Size}  & \multicolumn{7}{c|}{10k}                              & \multicolumn{2}{c|}{2k} & \multicolumn{2}{c|}{40k}        & \multicolumn{2}{c}{10k}       \\ 
\Xhline{2\arrayrulewidth}
\end{tabular}}
\caption{\textbf{Datasets and Tasks}. We use $3$ datasets with total $13$ tasks that covers various languages with different scales of training data.}
    \label{tab:datasets} \vspace{5mm}
\end{table*}
\begin{table*}[ht]
\centering
\resizebox{0.95\linewidth}{!}{
\begin{tabular}{lcccccccc}
\Xhline{2\arrayrulewidth}
\multicolumn{1}{l|}{\textbf{Dataset}}              & \multicolumn{3}{c|}{Multi30K}                                                           & \multicolumn{3}{c|}{WIT}                                                                                      & \multicolumn{2}{c}{WMT}                               \\ \hline
\multicolumn{1}{l|}{\textbf{Task}}                 & \multicolumn{3}{c|}{All}                                                                & \multicolumn{1}{c|}{Well-Res.} & \multicolumn{1}{c|}{Non-English} & \multicolumn{1}{c|}{Under-Res.} & \multicolumn{1}{c|}{Well-Res.} & Under-Res. \\ \hline
\multicolumn{1}{l|}{\textbf{Model}}                & \multicolumn{1}{c|}{Base}   & \multicolumn{1}{c|}{Small}  & \multicolumn{1}{c|}{Tiny}   & \multicolumn{1}{c|}{Base}           & \multicolumn{2}{c|}{Small}                                              & \multicolumn{1}{c|}{Base}           & Small           \\ \Xhline{2\arrayrulewidth}
\multicolumn{9}{c}{\textit{Architecture}}                                                                                                                                                                                                                                                                            \\ \Xhline{2\arrayrulewidth}
\multicolumn{1}{l|}{\textbf{Enc./Dec. Layers}}     & \multicolumn{1}{c|}{6}      & \multicolumn{1}{c|}{4}      & \multicolumn{1}{c|}{4}      & \multicolumn{1}{c|}{6}              & \multicolumn{2}{c|}{4}                                                  & \multicolumn{1}{c|}{6}              & 3               \\ \hline
\multicolumn{1}{l|}{\textbf{Embedding Dim.}}       & \multicolumn{1}{c|}{512}    & \multicolumn{1}{c|}{256}    & \multicolumn{1}{c|}{128}    & \multicolumn{1}{c|}{512}            & \multicolumn{2}{c|}{256}                                                & \multicolumn{1}{c|}{512}            & 512             \\ \hline
\multicolumn{1}{l|}{\textbf{Feedforward Dim.}}     & \multicolumn{1}{c|}{2048}   & \multicolumn{1}{c|}{256}    & \multicolumn{1}{c|}{256}    & \multicolumn{1}{c|}{2048}           & \multicolumn{2}{c|}{256}                                                & \multicolumn{1}{c|}{2048}           & 1024            \\ \hline
\multicolumn{1}{l|}{\textbf{Attn. Heads}}          & \multicolumn{1}{c|}{8}      & \multicolumn{1}{c|}{8}      & \multicolumn{1}{c|}{4}      & \multicolumn{1}{c|}{8}              & \multicolumn{2}{c|}{8}                                                  & \multicolumn{1}{c|}{8}              & 8               \\ \Xhline{2\arrayrulewidth}
\multicolumn{9}{c}{\textit{Optimization}}                                                                                                                                                                                                                                                                            \\ \Xhline{2\arrayrulewidth}
\multicolumn{1}{l|}{\textbf{Iters. / Warm-up}} & \multicolumn{3}{c|}{20k   / 2k}                                                         & \multicolumn{3}{c|}{50k   / 8k}                                                                               & \multicolumn{1}{c|}{150k   / 8k}    & 40k   / 4k      \\ \hline
\multicolumn{1}{l|}{\textbf{Batch Size (Tokens)}}  & \multicolumn{3}{c|}{2048}                                                               & \multicolumn{3}{c|}{4096}                                                                                     & \multicolumn{1}{c|}{16384}          & 8192            \\ \hline
\multicolumn{1}{l|}{\textbf{Learning Rate}}        & \multicolumn{1}{c|}{0.0001} & \multicolumn{1}{c|}{0.0005} & \multicolumn{1}{c|}{0.0025} & \multicolumn{3}{c|}{0.0005}                                                                                   & \multicolumn{1}{c|}{0.0005}         & 0.001           \\ \hline
\multicolumn{1}{l|}{\textbf{Dropout}}              & \multicolumn{1}{c|}{0.5}    & \multicolumn{1}{c|}{0.5}    & \multicolumn{1}{c|}{0.3}    & \multicolumn{1}{c|}{0.3}            & \multicolumn{1}{c|}{0.3}         & \multicolumn{1}{c|}{0.5}             & \multicolumn{1}{c|}{0.1}            & 0.3             \\ 
\Xhline{2\arrayrulewidth}
\end{tabular}
}
\caption{\textbf{Model Architectures and Optimization Hyperparameters.} Hyperparameters are selected by grid search on the respective validation set. Note that our \emph{Small} model in Multi30K is different from that used by Wu et al.~\cite{wu2021good}.}
\label{tab:arch_optim} \vspace{5mm}
\end{table*}

\subsection{Data Preprocessing}

Table~\ref{tab:datasets} summarizes the list of all machine translation tasks. We use byte-pair encoding (BPE)~\cite{gage1994new,sennrich2015neural} to tokenize all source and target sentences\footnote{\url{https://github.com/rsennrich/subword-nmt}}, with vocabulary size provided in the last row of the table. All sentences are preprocessed and cleaned using standard scripts\footnote{\url{https://github.com/moses-smt/mosesdecoder}}. 

\vspace{1mm}
\noindent\textbf{Multi30K.} This is a multilingual translation dataset with 29000 training samples of images and their annotations in English, German, French and Czech. Each English description is manually translated to German by a professional translator, then expanded to French and Czech. 
We use English-German (EN$\rightarrow$DE) and English-French (EN$\rightarrow$FR) for our experiments. Besides showing results on Test2016 and Test2017 sets, we use MSCOCO for evaluation which is a small dataset collected in WMT2017 multimodal machine translation challenge for testing out-of-domain performance of translation models. This evaluation set includes 461 more challenging out-of-domain instances with ambiguous verbs.

\vspace{1mm}
\noindent\textbf{WIT.} We construct multimodal translation datasets for 7 language pairs from WIT~\cite{srinivasan2021wit} data. 
Sentence-image data for MMT are obtained from \emph{reference descriptions} of the dataset, i.e., captions which are visible on the wiki page directly below the image. We empirically find these to contain richest visually grounded concepts compared to other types of captions provided in WIT.
First, we generate raw ground-truth translation pairs by sampling from images with captions annotated in both source and target languages. For images associated with multiple captions in the same language, we sample one sentence at random. Finally, a cleaning process filters out noise by ranking the sentence pairs by their length ratios $S/T$, and discarding top and bottom $5\%$ samples.

The validation and test splits for the original WIT are not publicly available, so we partition the training data to construct new splits for WIT with sizes provided in table~\ref{tab:datasets}.

\vspace{1mm}
\noindent\textbf{WMT.} We use the official train, validation and test data for standard WMT tasks. Under-resourced variants are created by downsampling the training sets of EN$\rightarrow$DE and EN$\rightarrow$FR tasks by approximately $3 \times 10^{-2}$ and $3 \times 10^{-3}$ respectively, creating subsets of 100k samples each. Validation and test sets kept the same as full WMT.

\vspace{1mm}
\noindent\textbf{Images.} Discrete visual encoders (VQGAN VAE) are trained on images randomly cropped and resized to $128 \time 128$ pixels, with pixel values rescaled to $[0, 1]$. At test time, we use center cropping for all images instead.

\subsection{Licenses}

% \texttt{\footnotesize State the name of the license (e.g., CC-BY 4.0) for each asset. If you scraped data from a particular source (e.g., website), you should state the copyright and terms of service of that source. If you are repackaging an existing dataset, you should state the original license as well as the one for the derived asset (if it has changed).}

All the datasets considered in this work are publicly available. WIT\footnote{\url{https://github.com/google-research-datasets/wit}}~\cite{srinivasan2021wit} is available under the CC BY-SA 3.0 license. Licenses for WMT 2014\footnote{\url{https://www.statmt.org/wmt14/translation-task.html}}~\cite{bojar2014findings} and Multi30K\footnote{\url{https://github.com/multi30k/dataset}}~\cite{elliott2016multi30k} are unknown. Use of images from Flickr30K\footnote{\url{http://hockenmaier.cs.illinois.edu/DenotationGraph/}}~\cite{young2014image} are subject to Flickr terms of use\footnote{\url{https://www.flickr.com/help/terms/}}.
\section{Implementation Details}

In  this  section,  we  provide  more  implementation  details regarding model architectures, training, inference procedures and hyperparameter selections. 

\def\textonly{\texttt{T}}
\def\halluc{\texttt{V}}
\def\hallucgt{\texttt{VM}}

\begin{table*}[ht]
\centering
\resizebox{\linewidth}{!}{
\begin{tabular}{@{}l|c|c|ccc|c|ccc|c@{}}
\Xhline{2\arrayrulewidth} 
\multirow{2}{*}{\textbf{Method}} &
\multirow{2}{*}{\textbf{Model}} & 
\multirow{2}{*}{\textbf{Params}} & \multicolumn{4}{c|}{\textbf{EN $\rightarrow$ DE}} & \multicolumn{4}{c}{\textbf{EN $\rightarrow$ FR}} \\
\cline{4-11}
& & & \textbf{Test2016} & \textbf{Test2017} & \textbf{MSCOCO} & \textbf{Average} & \textbf{Test2016} & \textbf{Test2017} & \textbf{MSCOCO} & \textbf{Average} \\
 \Xhline{2\arrayrulewidth} 
Transformer-Base & \textonly  & 49.1M & 61.8 ± 1.3 & 53.3 ± 1.1 & 49.1 ± 1.2   &  54.7 ± 1.2      & 80.1 ± 0.3
          & 74.5 ± 0.3         & 68.5 ± 0.2  & 74.4 ± 0.3   \\ \hline
\multirow{3}{*}{Transformer-Small} & \textonly & 9.2M & 65.6 ± 0.3 & 58.1 ± 0.6         
                & 52.5 ± 0.7   & 58.7 ± 0.5   & 79.2 ± 0.2         
                & 73.7 ± 0.1          & 67.9 ± 0.2  & 73.6 ± 0.2     \\
            & \halluc  & 24.3M    & \textbf{66.7 ± 0.4}         & \textbf{60.1 ± 0.0}
                    & \textbf{54.2 ± 0.4}   &  \textbf{60.3 ± 0.3}    & \textbf{80.3 ± 0.2}
                    & \textbf{74.8 ± 0.6} & 68.8 ± 0.4 & 74.6 ± 0.4 \\
            & \hallucgt  & 24.3M  & \textbf{66.7 ± 0.3} & \textbf{60.1 ± 0.0}
                    & \textbf{54.2 ± 0.3} & \textbf{60.3 ± 0.2} & \textbf{80.3 ± 0.2}
                    & 74.7 ± 0.5 & \textbf{69.0 ± 0.3} & \textbf{74.7 ± 0.3} \\ \hline
\multirow{3}{*}{Transformer-Tiny}  & \textonly  & 2.6M & 67.8 ± 0.3  & 61.6 ± 0.5
                    & 56.2 ± 0.6   & 61.9 ± 0.5    & 80.6 ± 0.2
                    & 75.6 ± 0.2          & 69.8 ± 0.2   & 75.3 ± 0.2     \\
            & \halluc  & 22.1M    & \textbf{68.8 ± 0.2}         & \textbf{62.5 ± 0.2}
            & 57.0 ± 0.6 & \textbf{62.8 ± 0.3} & \textbf{81.4 ± 0.2}
            & \textbf{76.4 ± 0.2}          & 70.9 ± 0.3  & 76.2 ± 0.2     \\
            & \hallucgt  & 22.1M    & 68.7 ± 0.2 & \textbf{62.5 ± 0.2}
                    & \textbf{57.2 ± 0.7}    &  \textbf{62.8 ± 0.4}  & \textbf{81.4 ± 0.2}
                    & \textbf{76.4 ± 0.1} & \textbf{71.0 ± 0.3} & \textbf{76.3 ± 0.2} \\
\Xhline{2\arrayrulewidth}
\end{tabular}}
\caption{\textbf{METEOR score on Multi30K}. \textonly: Baseline text-only transformer; \halluc: \ours model with hallucinated visual representations; \hallucgt: \ours model with ground-truth visual representations. 
Similar to BLEU score, \ours (\texttt{V}) consistently outperforms the text-only baseline while being very competitive with \ours (\texttt{VM}) on both tasks.
}
\label{tab:m30k_meteor}
\end{table*}
\begin{table*}[ht]
\small
\centering

\resizebox{0.9\linewidth}{!}{
\begin{tabular}{@{}l|ccc|cc|cc|c@{}}
\Xhline{2\arrayrulewidth}
\multirow{2}{*}{\textbf{Method}} & \multicolumn{3}{c|}{\textbf{Well-Resourced}}    & \multicolumn{2}{c|}{\textbf{Non-English}}  & \multicolumn{2}{c|}{\textbf{Under-Resourced}} & \multirow{2}{*}{\textbf{Average}}    \\
\cline{2-8} 
 & \textbf{EN $\rightarrow$ DE}        & \textbf{EN $\rightarrow$ ES}     & \textbf{EN $\rightarrow$ FR}        & \textbf{DE $\rightarrow$ ES}    & \textbf{ES $\rightarrow$ FR}        & \textbf{EN $\rightarrow$ RO} & \textbf{EN $\rightarrow$ AF} &    \\ \Xhline{2\arrayrulewidth}
Text-Only   & 35.4 ± 0.5          & 44.6 ± 1.7          & 37.4 ± 1.3    & 33.3 ± 0.3  & 37.0 ± 0.2  & 26.6 ± 0.6 & 30.2 ± 1.0   & 34.9 ± 0.8  \\
UVR-NMT~\cite{zhang2020neural}   &  35.9 ± 0.1 & 46.7 ± 0.2          &    39.5 ± 0.5       &  32.7 ± 1.1         &  37.2 ± 0.7        &  28.0 ± 0.7         & 32.8 ± 1.4           & 36.1 ± 0.7  \\ 
RMMT~\cite{wu2021good}    & 35.4 ± 0.6          & 44.8 ± 0.8        & 39.0 ± 1.0         & 33.2 ± 0.4          & 36.5 ± 0.9         & 23.6 ± 0.2 & 29.6 ± 1.3         & 34.6 ± 0.7 \\
\ours   & \textbf{36.8 ± 0.5}   & \textbf{47.1 ± 0.2}         & \textbf{40.2 ± 0.3}          & \textbf{34.3 ± 0.3}        & \textbf{37.5 ± 0.9}          & 30.4 ± 0.9 & \textbf{34.2 ± 0.2}          &  \textbf{37.2 ± 0.5}  \\
\oursm  & 36.7 ± 0.5  & \textbf{47.1 ± 0.3}  & \textbf{40.2 ± 0.3}          & \textbf{34.3 ± 0.4}         & \textbf{37.5 ± 0.9}          & \textbf{30.5 ± 0.9} & \textbf{34.2 ± 0.2}          & \textbf{37.2 ± 0.5} \\
\Xhline{2\arrayrulewidth}
\end{tabular}}

\caption{\textbf{METEOR score on WIT}. Our proposed, \ours achieves an average $4$ point improvement over text-only baseline in under-resource setting including best average performance among all compared methods.}
\label{tab:wit_meteor}
\end{table*}

\subsection{Model Architecture}
In Table~\ref{tab:arch_optim} we provide the detailed architectures of all translation models $\mathbf{f_T}$ used for each dataset. For all experiments, we use a hallucination transformer $\mathbf{f_H}$ of depth 2 and VQGAN VAE visual encoder $\mathbf{f_V}$ of encoder depth 6.

Sinusoidal positional embeddings (PE)~\cite{vaswani2017attention} are added to the multimodal input sequence to the translation transformer $\mathbf{f_T}$. Using a learnable PE~\cite{gehring2017convolutional} did not improve translation performance in preliminary experiments. For visual tokens, we follow \cite{ramesh2021zero} to compute 2D positional encoding as the sum of row and column embeddings.

\subsection{Training Procedure}
For all models and tasks, we optimize the \ours system using Adam~\cite{kingma2014adam} with inverse square root learning rate schedule and warm-up steps. Table~\ref{tab:arch_optim} lists important optimization hyperparameters used for each task and model, determined in preliminary experiments by grid search on the respective validation set.
% We use xx NVIDIA Tesla V100 GPUs for experiments on MMT and WIT and 18 GPUs for both RGB + Flow and RGB + Flow + Audio experiments

\subsection{Inference and Evaluation}
During inference we use beam search with a beam size of 5 to generate translation outputs for each task. Length penalty $\alpha$ is set to 0.6 on full WMT dataset, 2 on WIT dataset, and 1 on all other trasnlation tasks. We use standard scripts to compute BLEU\footnote{\url{https://github.com/moses-smt/mosesdecoder/blob/RELEASE-4.0/scripts/generic/multi-bleu.perl}}~\cite{papineni2002bleu} 
and METEOR\footnote{\url{https://github.com/cmu-mtlab/meteor}}~\cite{denkowski2014meteor} 
scores as evaluation metrics for machine translation. 

\subsection{Code and Models}
Our \ours framework is implemented on top of fairseq~\cite{ott2019fairseq} using PyTorch~\cite{paszke2019pytorch}. 
% Please refer to \textbf{\enquote{VALHALLA\_code.zip}} in the supplementary material for our code submission. We will make the code publicly available after the acceptance.
Code and pretrained models are available at our project page: \url{http://www.svcl.ucsd.edu/projects/valhalla}.
\begin{figure*}[h!]
    \centering
    \includegraphics[width=0.23\linewidth]{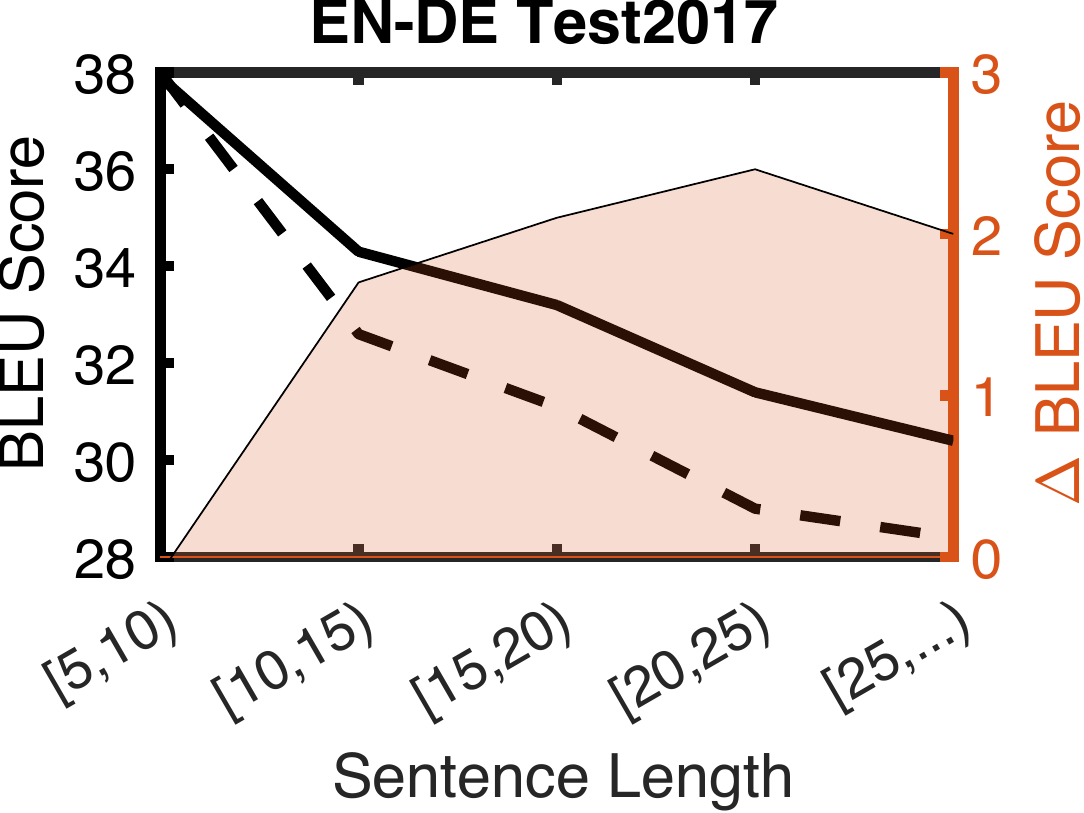} \quad
    \includegraphics[width=0.23\linewidth]{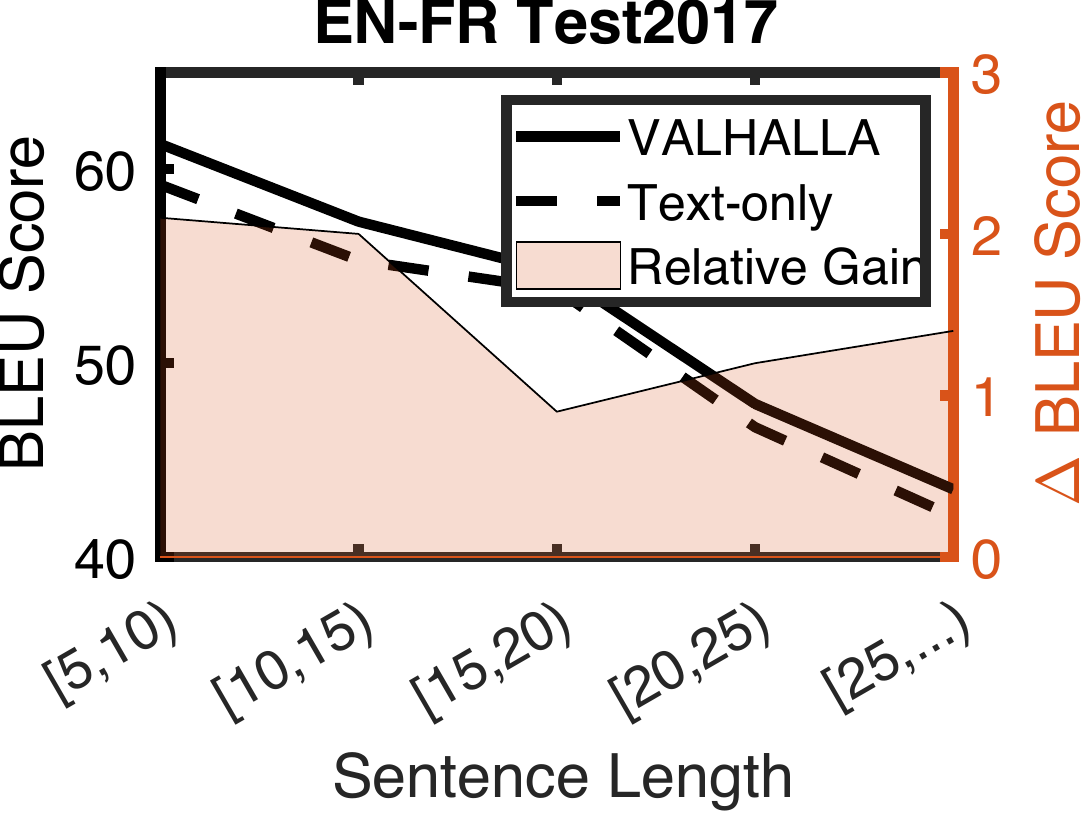} \quad
    \includegraphics[width=0.23\linewidth]{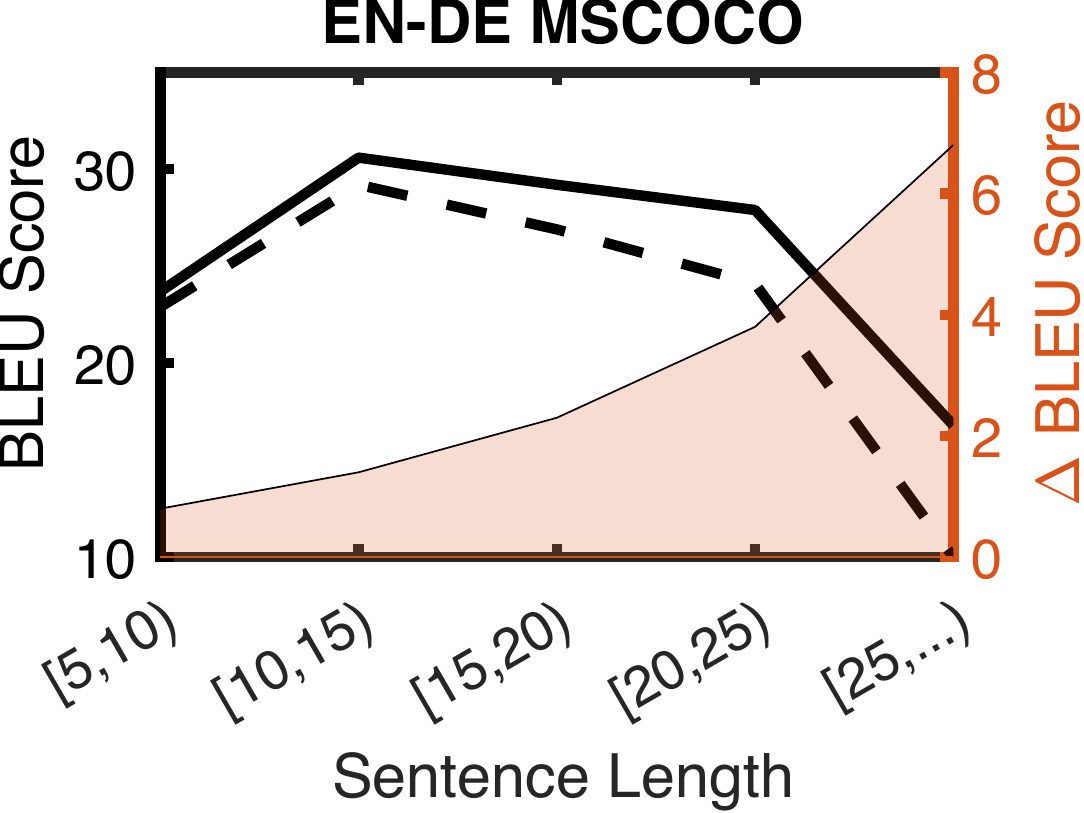} \quad
    \includegraphics[width=0.23\linewidth]{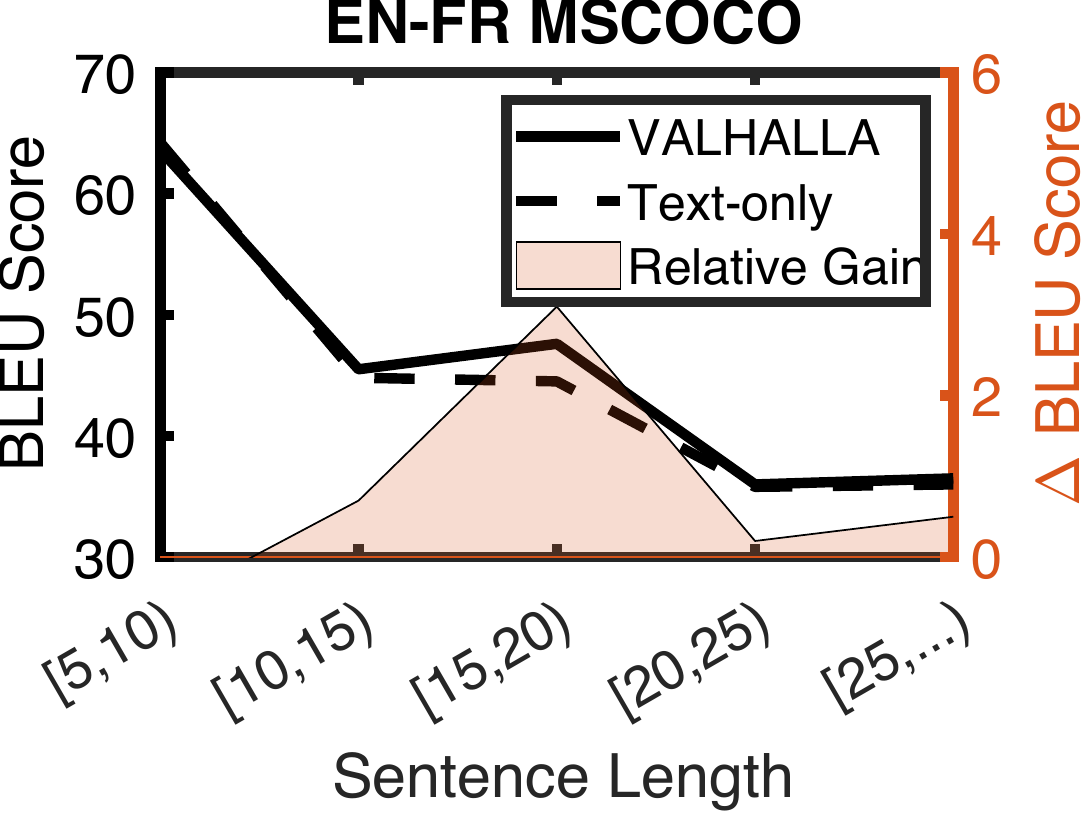}
    \caption{\textbf{Performance vs. Sentence Length.} We report BLEU scores on different groups divided according to source sentence lengths on Multi30K Test2017 and MSCOCO split.}
    \label{fig:input_length_supp}
\end{figure*}
\begin{figure*}[h!]
    \centering
    \includegraphics[width=0.23\linewidth]{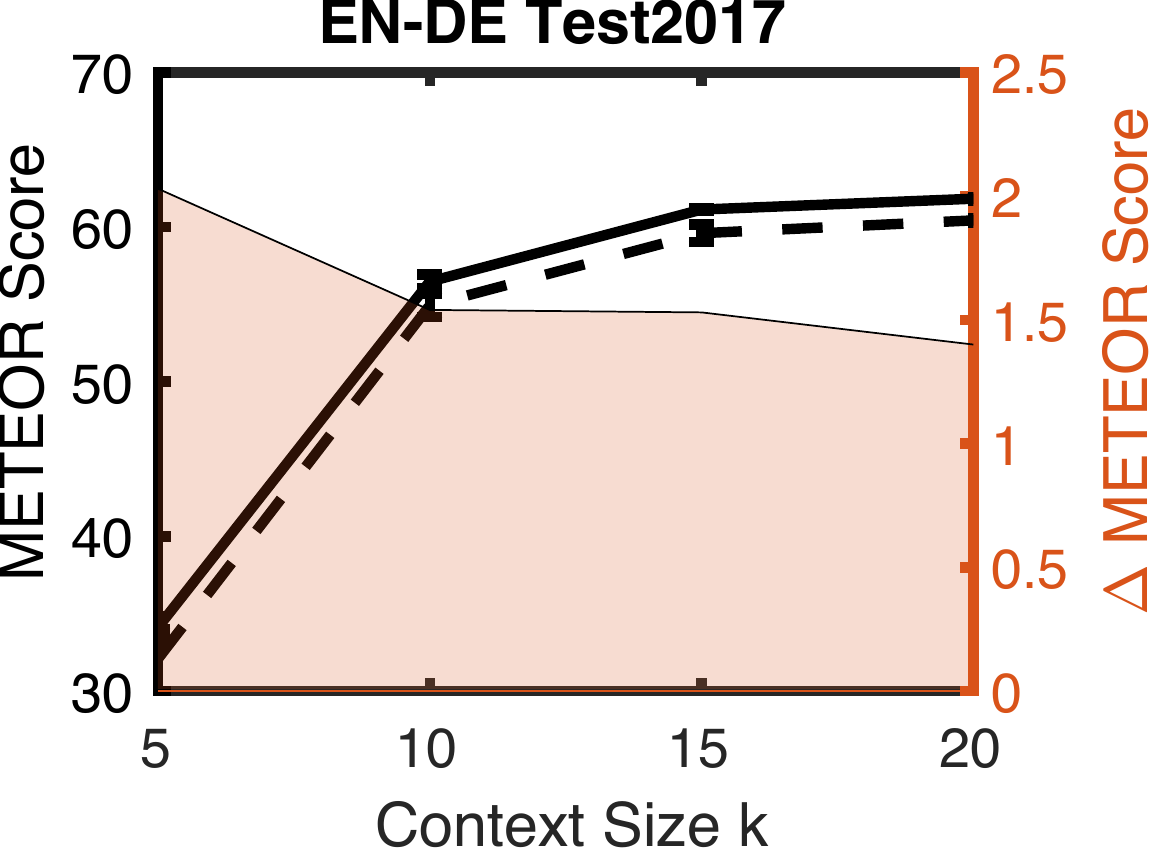} \quad
    \includegraphics[width=0.23\linewidth]{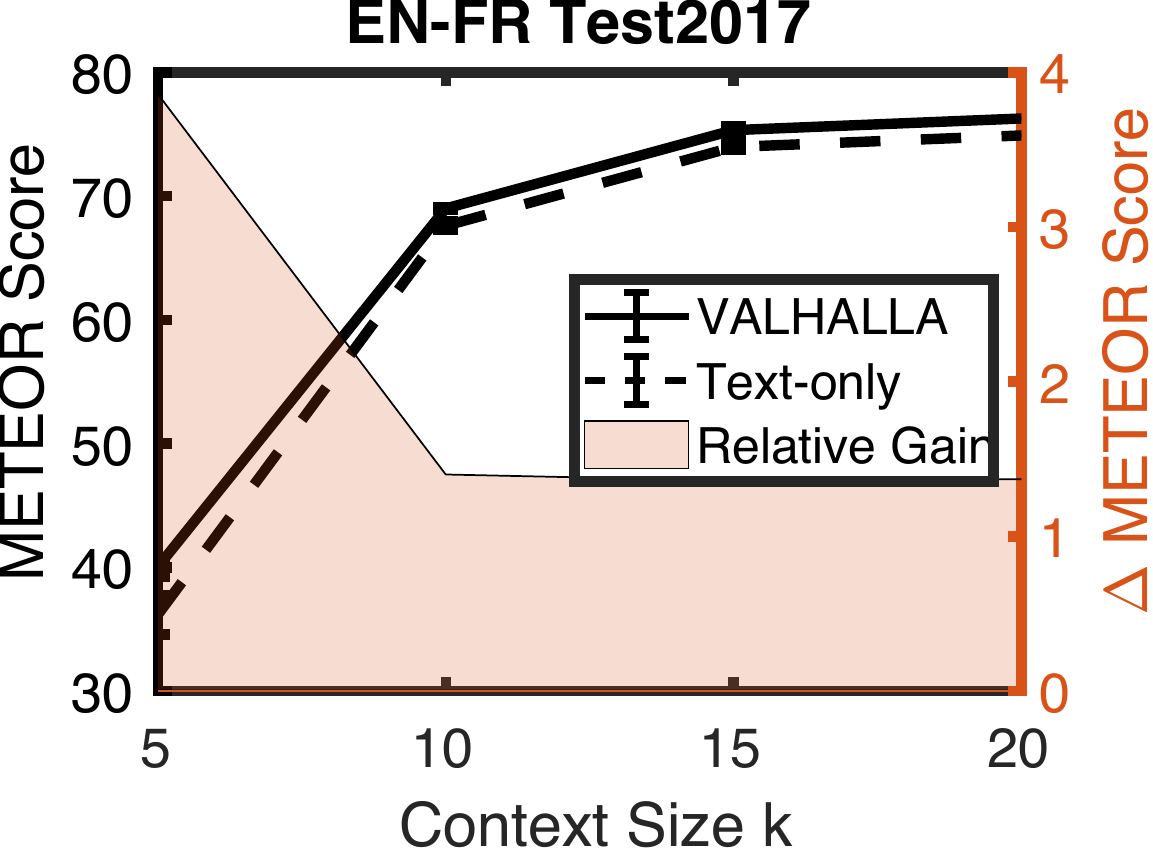} \quad
    \includegraphics[width=0.23\linewidth]{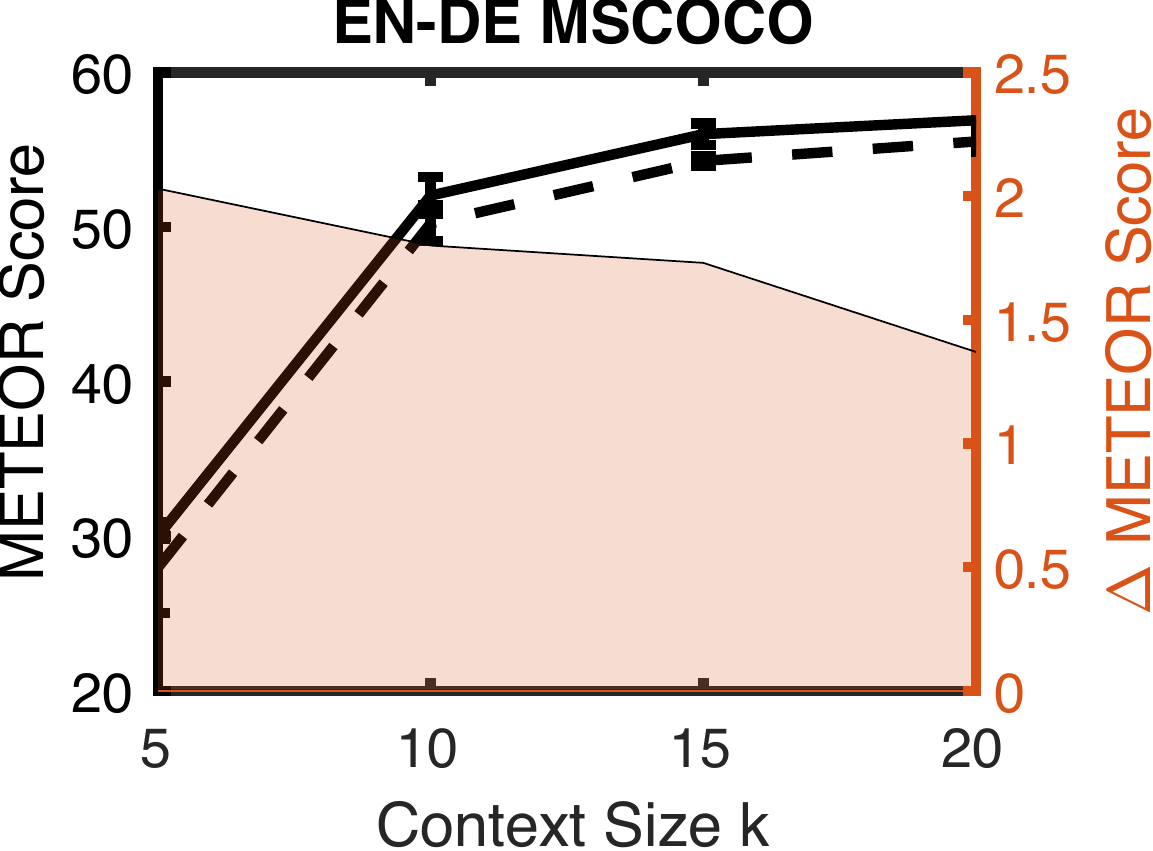} \quad
    \includegraphics[width=0.23\linewidth]{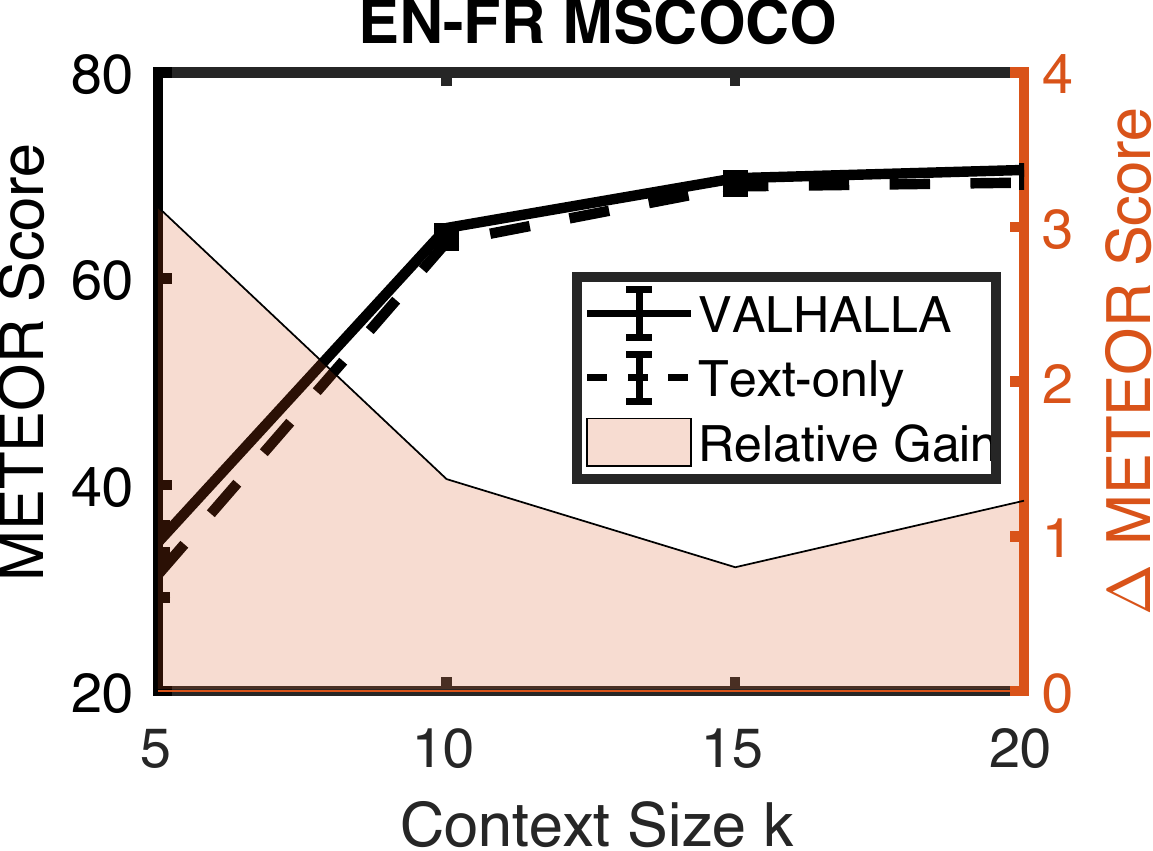}
    \caption{\textbf{Evaluation with Progressive Masking.} We plot the METEOR scores of \ours and text-only models evaluated on Test2017 and MSCOCO splits, as well as the relative improvements over the text-only baseline on both EN$\rightarrow$DE and EN$\rightarrow$FR tasks.}
    \label{fig:progressive_mask_supp}
\end{figure*}
\newcommand{\xmark}{\ding{55}}%

\begin{table*}[ht]
\centering
\resizebox{\linewidth}{!}{
\begin{tabular}[b]{@{}lccccccccc@{}}
\toprule
\textbf{Method} & \textbf{External Data} & \textbf{EN$\rightarrow$DE}        & \textbf{EN$\rightarrow$ES}     & \textbf{EN$\rightarrow$FR}        & \textbf{DE$\rightarrow$ES}    & \textbf{ES$\rightarrow$FR}        & \textbf{EN$\rightarrow$RO} & \textbf{EN$\rightarrow$AF} & \textbf{Average}   \\ \midrule
\multirow{2}{*}{\ours}  & \xmark & 17.5 ± 0.4   & 27.5 ± 0.2         & 18.8 ± 0.2          & \textbf{11.3 ± 0.2}        & 16.6 ± 0.8          & \textbf{14.4 ± 1.0} & \textbf{14.0 ± 0.5}          & 17.2 ± 0.4 \\
  & \checkmark & \textbf{18.0 ± 0.3}   & \textbf{27.7 ± 0.4}         & \textbf{19.1 ± 0.3}          & \textbf{11.3 ± 0.7}        & \textbf{17.4 ± 0.4}          & 14.1 ± 0.3 & 13.8 ± 0.3          & \textbf{17.3 ± 0.4} \\ \midrule
\multirow{2}{*}{\oursm} & \xmark & 17.4 ± 0.4  & 27.5 ± 0.2  & 18.8 ± 0.2          & \textbf{11.3 ± 0.2}         & 16.6 ± 0.8          & \textbf{14.4 ± 1.0} & \textbf{14.0 ± 0.4}          & 17.2 ± 0.4 \\
  & \checkmark & \textbf{18.0 ± 0.3}   & \textbf{27.8 ± 0.4}         & \textbf{19.1 ± 0.3}          & \textbf{11.3 ± 0.7}        & \textbf{17.4 ± 0.4}          & 13.9 ± 0.4 & 13.9 ± 0.4          & \textbf{17.3 ± 0.4} \\ 
\bottomrule
\end{tabular}}
\caption{\textbf{Training Visual Encoder (VQGAN VAE) with External Data on WIT.}}
\label{tab:ablations_external}
\end{table*}
\begin{table*}[ht]
\centering
\resizebox{0.85\linewidth}{!}{
\begin{tabular}{@{}l|ccc|c|ccc|c@{}}
\Xhline{2\arrayrulewidth} 
\multirow{2}{*}{\textbf{Method}} &\multicolumn{4}{c|}{\textbf{EN $\rightarrow$ DE}} & \multicolumn{4}{c}{\textbf{EN $\rightarrow$ FR}} \\
\cline{2-9}
& \textbf{Test2016} & \textbf{Test2017} & \textbf{MSCOCO} & \textbf{Average} & \textbf{Test2016} & \textbf{Test2017} & \textbf{MSCOCO} & \textbf{Average} \\
 \Xhline{2\arrayrulewidth} 
Text-Only & 38.2 ± 0.4 & 28.8 ± 0.4         
                & 25.8 ± 0.3   & 30.9 ± 0.4    & 58.4 ± 0.4         
                & 50.9 ± 0.3          & 41.6 ± 0.4  & 50.3 ± 0.4     \\
CLIP & 38.7 ± 0.2 & 30.1 ± 0.3         
                & 27.3 ± 0.6   & 32.1 ± 0.3    & 59.0 ± 0.6         
                & 51.6 ± 0.3          & 42.6 ± 0.6  & 51.1 ± 0.5     \\
\ours     & \textbf{39.4 ± 0.3}         & \textbf{31.7 ± 0.2}
                    & \textbf{27.9 ± 0.3}   &  \textbf{33.0 ± 0.3}     & \textbf{60.5 ± 0.1}
                    & \textbf{52.3 ± 0.7} & \textbf{43.1 ± 0.3} & \textbf{52.0 ± 0.4} \\
\Xhline{2\arrayrulewidth}
\end{tabular}}
\caption{\textbf{BLEU score with CLIP Hallucination}, evaluated with Transformer-Small models on Multi30K.} \vspace{-2mm}
\label{tab:clip_halluc}
\end{table*}

\section{Additional Results}

\subsection{Numerical Scores}

\vspace{1mm}
\noindent\textbf{METEOR Scores.} Tables~\ref{tab:m30k_meteor} and \ref{tab:wit_meteor} summarizes METEOR scores of models evaluated on Multi30K~\cite{elliott2016multi30k} and WIT~\cite{srinivasan2021wit}, respectively. Similar to the trend in BLEU scores (Tables 1 and 3 in the main paper), \ours outperforms text-only and multimodal baselines consistently on all tasks.

\vspace{1mm}
\noindent\textbf{Sentence Length.} We repeat the study of translation performance vs. length of source sentence (Figure~\ref{fig:input_length}) on Test2017 and MSCOCO evaluation sets. Figure~\ref{fig:input_length_supp} shows the results.
Similar to the observations in Test2016 set, \ours generally produces larger gains over text-only baselines for longer sentences ($> 10$ source tokens).

\vspace{1mm}
\noindent\textbf{Progressive Masking.} Figure~\ref{fig:progressive_mask_supp} compares the performance of \ours and text-only model under progressive masking, evaluated on Test2017 and MSCOCO splits. Similar to the observations in Figure~\ref{fig:progressive_mask}, we observe a larger gap between \ours and text-only model with low context sizes $k$, validating its effectiveness in translating ambiguous or out-of-context sentences.

\vspace{1mm}
\noindent\textbf{Number of Parameters vs Performance.}
Larger model size does not guarantee stronger translation performance due to overfitting. As shown in Table~\ref{tab:m30k_ours}, among all three backbone architectures experimented on Multi30K, Transformer-Tiny with the least number of parameters achieved the highest scores, consistent with the findings of~\cite{wu2021good}. Our proposed, \ours achieves 10 BLEU points (8 METEOR points) higher than the text-only baseline transformer on Multi30K EN$\rightarrow$DE, and 3 BLEU points (2 METEOR points) higher on EN$\rightarrow$FR tasks, while using $2\times$ fewer parameters.

\subsection{Ablation Studies}

\vspace{1mm}
\noindent\textbf{Effect of External Data on WIT Tasks.} Table~\ref{tab:ablations_external} shows full results on WIT with a universal visual encoder pretrained on the union of images from all tasks. While this improves performance on 4 out of 7 tasks, average score over all tasks is only marginally better than individually trained encoders.

\vspace{1mm}
\noindent\textbf{CLIP Hallucination.} A naive method to predict visual features from an input sentence is to utilize a CLIP model~\cite{radford2021learning}, learned with a cross-modal contrastive loss that aligns the embedding space of text and image. We train a multimodal translation model with a gating strategy on top of image features extracted from the pretrained CLIP visual encoder, and replace this with text embeddings from CLIP language encoder to realize text-only inference. 
Table~\ref{tab:clip_halluc} shows the performance of this CLIP-based hallucination model on Multi30K. While CLIP-based feature hallucination consistently outperforms the text-only baseline, the improvements in BLEU score is not nearly as large as those achieved by \ours, which still reports the best results among all strategies. 

\vspace{1mm}
\noindent\textbf{Hallucinating Multiple Images.} We study the possibility of modifying the hallucination transformer $\mathbf{f_H}$ to predict multiple images for each input sentence. While this in theory enhances the diversity of hallucination, we did not observe significant improvement over baselines. 
By hallucinating 5 images per example, Transformer-Small model achieved 39.4 ± 0.3 and 60.4 ± 0.2 BLEU score on EN$\rightarrow$DE and EN$\rightarrow$FR tasks respectively, evaluated on Test2016 split. On Test2017 split, the scores are 31.8 ± 0.4 and 52.2 ± 0.1. 
Both results are comparable to the results reported in main paper, suggesting that a single hallucination per sample is adequate to capture diverse visual concepts in the input sentence. 
% \todo{Retrieval using k-nearest neighbour}

% \vspace{1mm}
% \noindent\textbf{Randomized visual tokens.}
% On Multi30K'16 EN$\rightarrow$FR, replacing inputs to MMT transformer $\bf f_T$ with \emph{random} visual tokens reduced BLEU score to 61.2 from 62.3. \todo{Results on other tasks} The consistent drop in performance by using randomized tokens suggests that visual hallucination under limited textual context is indeed of crucial significance.

\vspace{1mm}
\noindent\textbf{Pretrained VAEs.}
Using the pretrained VAE from DALL-E as the visual encoder gives poor results (58.8 BLEU on Multi30K'16 EN$\rightarrow$FR). We attribute this to the large visual sequence length (32$\times$32) used by DALL-E, which prevents the MMT transformer to attend to text tokens, as analyzed in Table~\ref{tab:ablations_encoder_size}.
Likewise, use of pre-trained VQGAN VAE~\cite{esser2021taming} with 16$\times$16 latent visual sequence also does not improve results from training on Multi30K alone (59.5 vs. 60.5 BLEU), likely due to the larger sequence length or domain gap between pretraining datasets. 

\subsection{Qualitative Examples}
\begin{figure*}[ht]
    \centering
    \begin{subfigure}[b]{\linewidth}
        % \centering
        \includegraphics[width=0.45\linewidth]{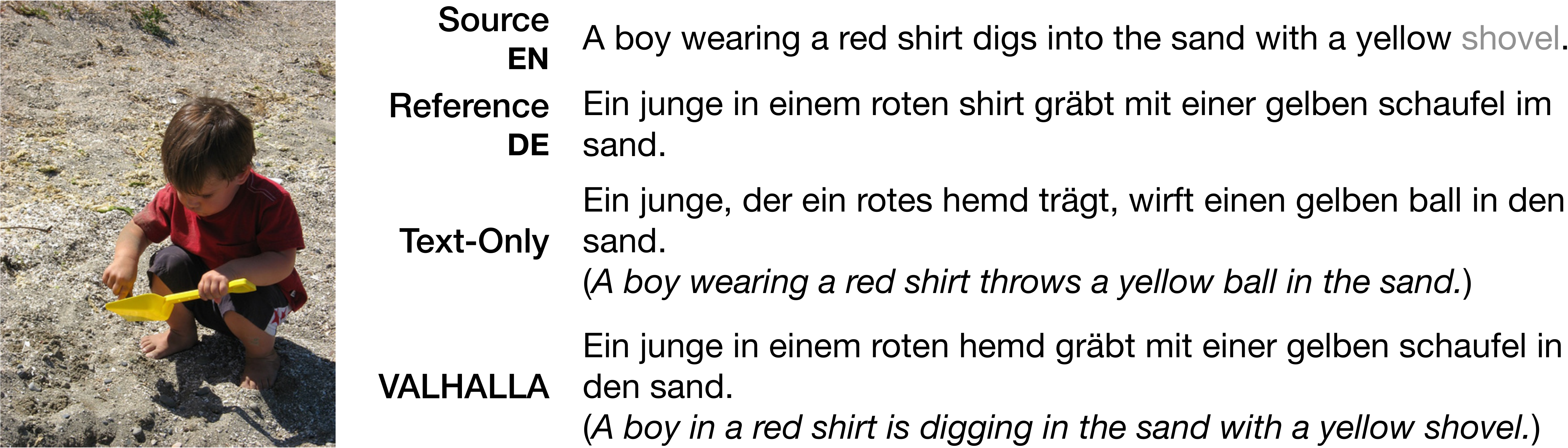} \qquad
        \includegraphics[width=0.45\linewidth]{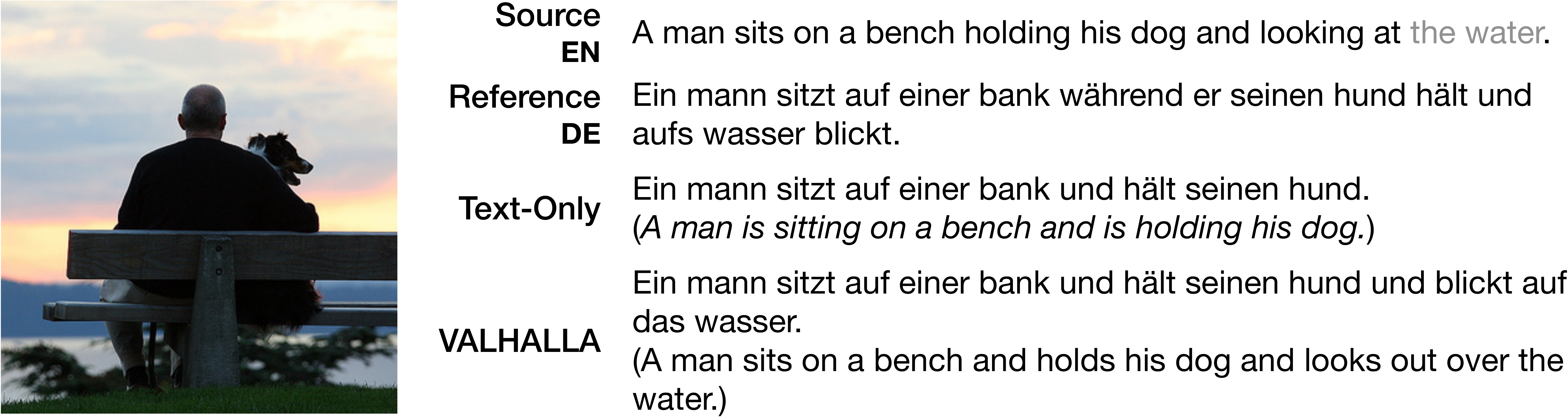} \\[3mm]
        \includegraphics[width=0.45\linewidth]{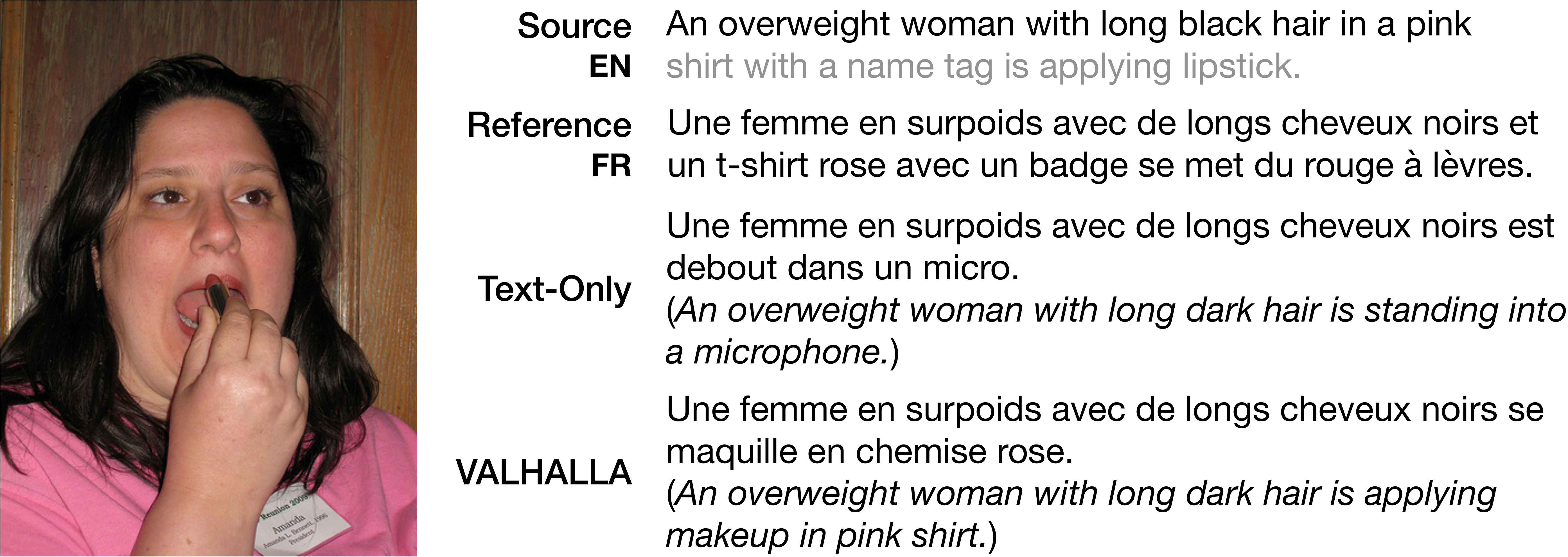} \qquad
        \includegraphics[width=0.45\linewidth]{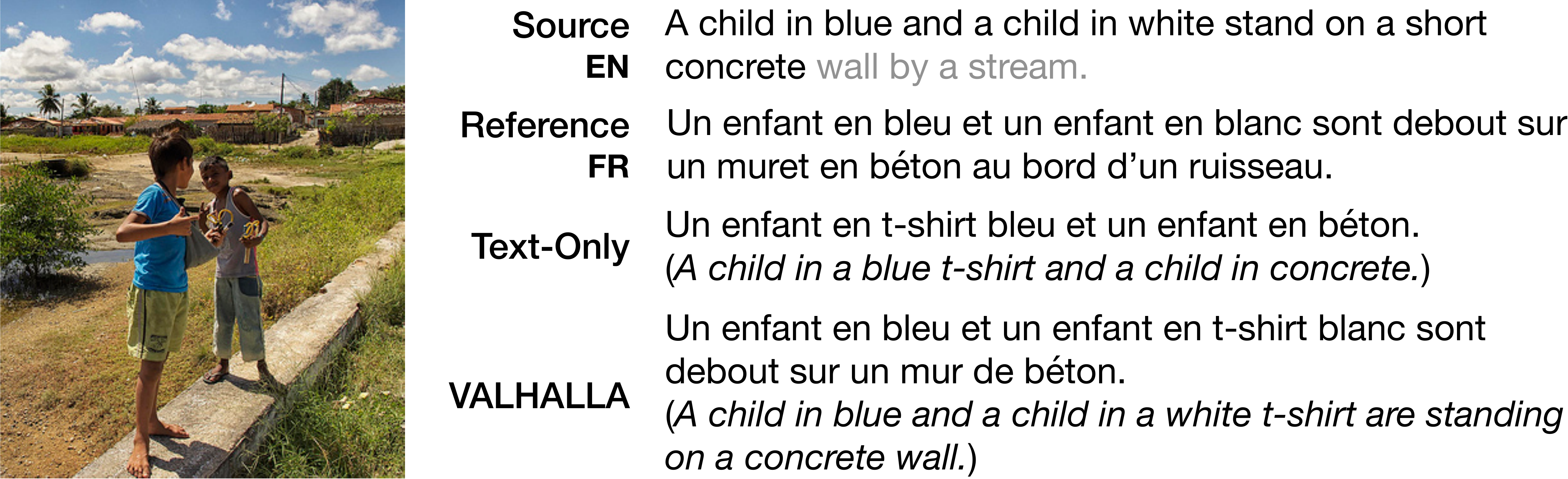}
        \caption{Progressive Masking.}
    \end{subfigure} \\[3mm]
    \begin{subfigure}[b]{\linewidth}
        % \centering
        \includegraphics[width=0.45\linewidth]{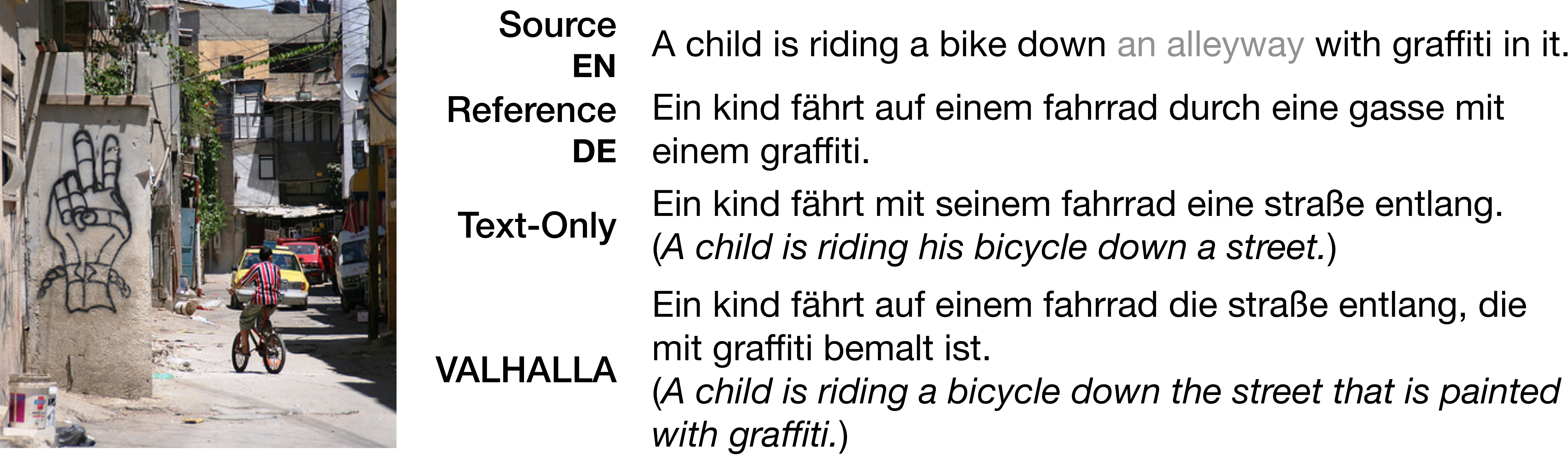} \qquad
        \includegraphics[width=0.45\linewidth]{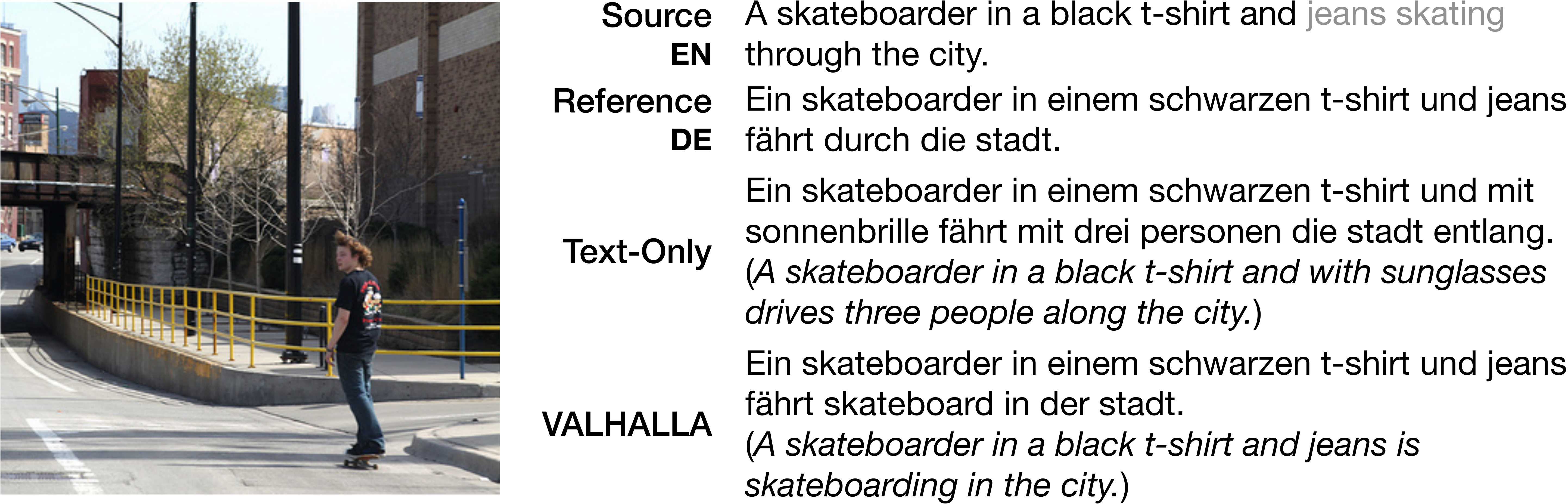} \\[3mm]
        \includegraphics[width=0.45\linewidth]{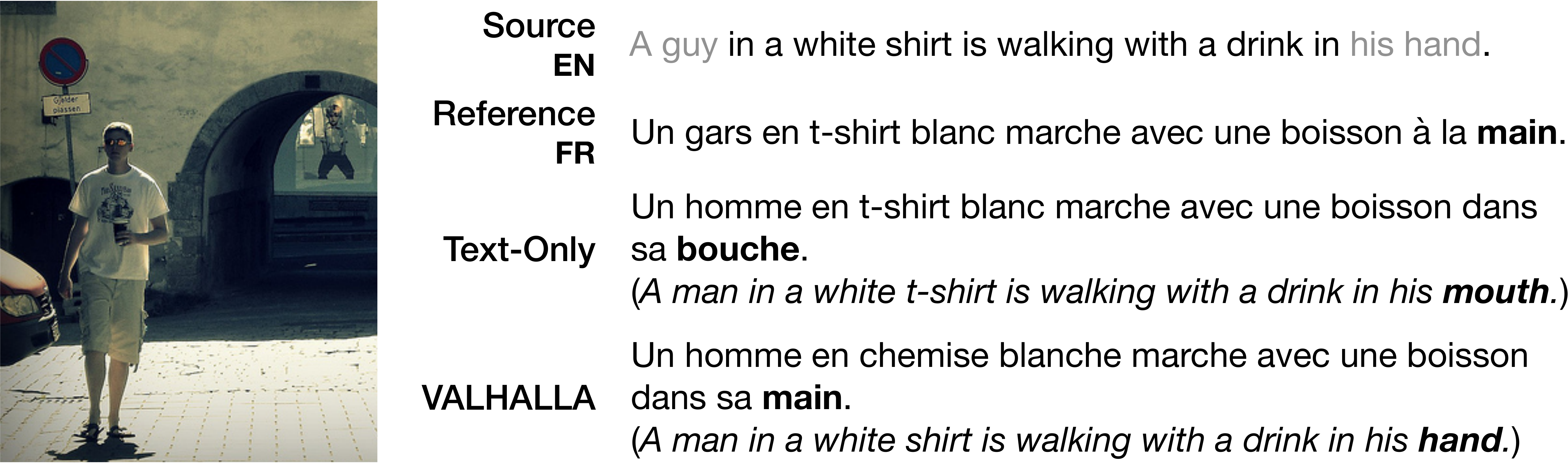}  \qquad
        \includegraphics[width=0.5\linewidth]{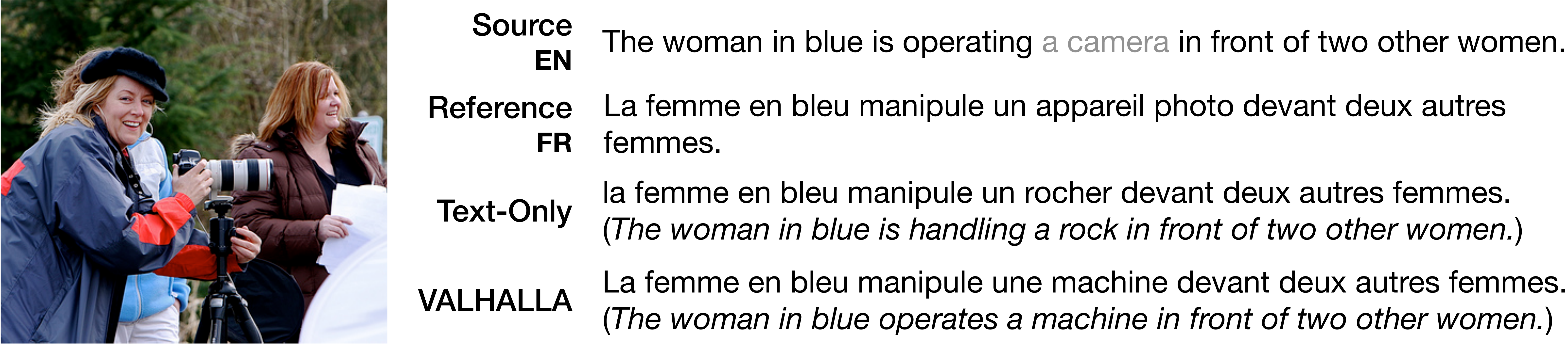}
        \caption{Visual Entity Masking.}
    \end{subfigure}
    \caption{\textbf{Qualitative Translation Results with Progressive Masking and Visual Entiry Masking.} Phrases in gray in the source sentence are masked with \texttt{<v>} at model input. \ours models generate more fluent and logical translations than text-only baseline transformer. Best viewed in color.
    }
    \label{fig:qualitative_supp}
\end{figure*}
\begin{figure*}[ht]
    \centering
    \includegraphics[width=0.7\linewidth]{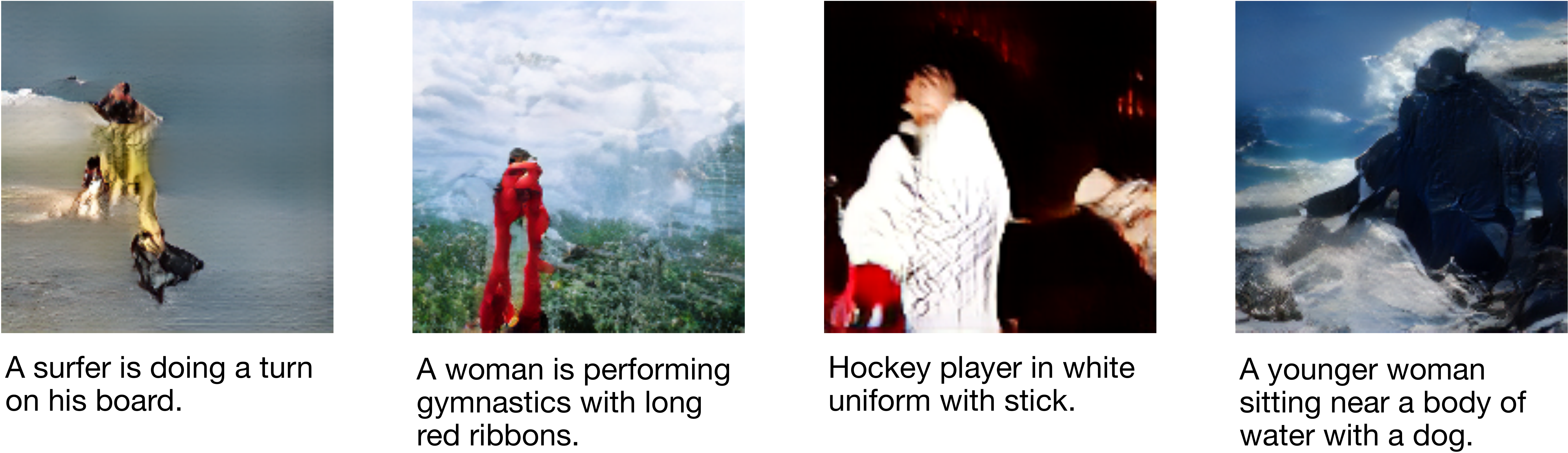}
    \caption{\textbf{Reconstruction of Hallucinated Visual Tokens.} We use the pretrained VQGAN VAE image decoder to visualize the hallucinated visual sequence (the image decoder is \emph{not} fine-tuned jointly with \ours). \ours captures abstract concepts such as ``surfer'' and ``red ribbons'', despite not being trained for high-quality image generation. Best viewed in color.}
    \label{fig:qualitative_halluc_supp}
\end{figure*}

\vspace{1mm}
\noindent\textbf{Translation under Limited Visual Context.} 
Figure~\ref{fig:qualitative_supp} shows additional qualitative translation results under both progressive masking and visual entity masking. We observe that in both EN$\rightarrow$DE and EN$\rightarrow$FR tasks 
% from Multi30K dataset
, our proposed \ours models are often capable of generating more fluent and logical translations than the text-only baseline transformer, by choosing plausible phrases to replace the masked tokens in the source sentences. 

% \vspace{1mm}
% \noindent\textbf{Attention Maps.} \todo{Results, discussions}

\vspace{1mm}
\noindent\textbf{Reconstructed Visual Hallucinations.}
Figure~\ref{fig:qualitative_halluc_supp} visualizes the hallucinated visual tokens using
the VQGAN VAE decoder, which is pretrained jointly with VAE encoder $\bf f_V$. As can be seen from the examples, our proposed \ours captures
abstract concepts such as ``surfer'' and ``red ribbons'', despite
not being trained for high-quality image generation.
\section{Limitations}

Effectiveness of our proposed approach depends on the availability of good quality images to train the visual hallucination transformer, which is often difficult to collect especially for languages beyond English. Another potential limitation is training complexity which we believe could be greatly improved if we pre-extract VQGAN-VAE tokens, like existing methods did with ResNet-based visual encoders. 
\section{Broader Impact}  

Our approach not only leads to more accurate translation systems on top of the existing text-only methods, but also breaks the major bottleneck of using visual information in multimodal machine translation. 
Our research can have a positive impact on many real-world applications of neural machine translation involving a broad range of languages. It improves  translation performance in both well- and under-resourced scenarios which is of great practical importance. 
% Our research can have a positive impact in improving the performance of MT systems in under-resourced scenarios which is of great practical importance. 
Negative impacts of our research are difficult to predict, however, it shares many of the pitfalls associated with standard MT models such as dataset/social bias and susceptibility to adversarial attacks. While we believe that these issues should be mitigated, they are beyond the scope of this paper.

}

\end{document}